\documentclass[nonacm,screen]{acmart}
\usepackage{csquotes}
\usepackage{float}
\usepackage{pifont}
\usepackage{adjustbox}
\newcommand{\cmark}{\ding{51}}%
\newcommand{\xmark}{\ding{55}}%
\usepackage{makecell}

\definecolor{darkblue}{rgb}{0, 0, 0.5}
\hypersetup{colorlinks=true,citecolor=darkblue, linkcolor=darkblue, urlcolor=darkblue}

\begin{document}
\title{A Survey On Neural Word Embeddings}

\author{Erhan Sezerer}
\email{erhansezerer@iyte.edu.tr}
\author{Selma Tekir}
\email{selmatekir@iyte.edu.tr}
\affiliation{%
  \institution{Izmir Institute of Technology}
  \city{Izmir}
  \state{Turkey}
  \postcode{35430}
}

\begin{abstract}
Understanding human language has been a sub-challenge on the way of intelligent machines. The study of meaning in natural language processing (NLP) relies on the distributional hypothesis where language elements get meaning from the words that co-occur within contexts. The revolutionary idea of distributed representation for a concept is close to the working of a human mind in that the meaning of a word is spread across several neurons, and a loss of activation will only slightly affect the memory retrieval process.

Neural word embeddings transformed the whole field of NLP by introducing substantial improvements in all NLP tasks. In this survey, we provide a comprehensive literature review on neural word embeddings. We give theoretical foundations and describe existing work by an interplay between word embeddings and language modeling. We provide broad coverage on neural word embeddings, including early word embeddings, embeddings targeting specific semantic relations, sense embeddings, morpheme embeddings, and finally, contextual representations. Finally, we describe benchmark datasets in word embeddings' performance evaluation and downstream tasks along with the performance results of/due to word embeddings.
\end{abstract}

\maketitle

\section{Introduction}
The recent decade has witnessed a transformation in natural language processing (NLP). This transformation can be attributed to neural language models, their success in representation learning, and the transfer of this knowledge into complex NLP tasks.

Before neural representation learning, representations of words or documents have been computed using the vector space model (VSM) of semantics. \citet{turney&pantel2010} provide a comprehensive survey on the use of VSM for semantics. In VSM \cite{salton1975}, frequencies of words in documents are considered to form a term-document matrix, and global co-occurrences of words in context lead to word-context matrices \cite{lsa, HAL, Hellinger-PCA}. Although these count-based representations are proved helpful in addressing semantics, they are the bag of words approaches and are not able to capture both syntactical and semantic features at the same time, which is required for performing well in NLP tasks.

Neural word embeddings are due to neural language models. Neural network architecture is constructed to predict the next word given the set of neighboring words in the sequence in neural language modeling. In the iterative processing of this prediction over a large corpus, the learned weights in the hidden layers serve as neural embeddings for words.

Neural word embeddings have experienced an evolution. Early word embeddings had some problems. Although they can learn syntactic and semantic regularities, they are not so good at capturing their mixture. Moreover, they provide just one representation that is shared among the different senses of a word. State-of-the-art contextual embeddings are responsive to these problems. They lead to a significant performance improvement and find their application throughout all NLP tasks and in many other fields \cite{app1, app2, app3}.

In this article, we describe this transition by first providing the theoretical foundations. Then, preliminary realizations of these ideas by some seminal papers are explained. In the remaining part, generally accepted and efficiently computable early word embeddings are introduced. Afterward, extensions to early word embeddings are given with respect to some criteria such as the use of knowledge base, having morphological features, and addressing specific semantic relations (synonym, antonym, hypernym, hyponym, etc.). Succeedingly, separate sections are devoted to sense, morphological, and contextual embeddings. We also include performance evaluation of word embeddings on the benchmark datasets. Finally, we conclude the article with some historical reflections and future remarks.
We have also included a diagram showing the milestone papers and summarizing the flow of ideas in the field in Appendix \ref{app-guide}.

Multilingual information requirements and parallel/comparable corpora in different languages pave the way for cross-lingual representations of words in a joint embedding space. In this survey, we exclude those techniques that specialize in learning word representations in a multilingual setting. The reader can refer to \citet{Ruder2019} for a comprehensive overview of cross-lingual word embedding models.

\section{Background}
\subsection{Distributional Hypothesis}
Together with \citet{Wittgenstein1953}, \citet{harris1954} were one of the first authors to propose that languages have a distributional structure. He argues that language elements are dispersed to environments that are composed of an existing array of their co-occurrents. An element's distribution is the sum of all these environments. Harris' second contribution is that we can relate an element's distribution with its meaning. He states that at least certain aspects of meaning are due to distributional relations. For instance, synonymy between two words can be defined as having almost identical environments except chiefly for glosses where they co-occur e.g. oculist and eye-doctor. The author also suggests that sentences starting with a pronoun should be considered as the same context as the previous sentence where the subject of the pronoun is given since their occurrence is not arbitrary and the fullest environmental unit for the distributional investigation is the connected discourse structures of such sentences.
\subsection{Distributional Representations}
\citet{Hinton1986} utilize the idea of distributed representations for concepts. They propose patterns of hidden layer activations (which are only allowed to be $0$ or $1$) as the representation of meanings. They argue that the most important evidence of distributed representations is their degree of similarity to the weaknesses and strengths of human mind. Unlike computer memory, human brain is able to retrieve memory from partial information. Distributed representations conform to this notion better than local distributions (i.e. bag of words model, where each meaning is associated with a single computational unit) since the meaning of a word is distributed across several units and a loss of an activation will only slightly effect the memory retrieval process. Rest of the activations that are still there will be able to retrieve the memory. Even if the occlusion of activations are strong enough to lead the system to an incorrect meaning, it will still result in a meaning close to that of the target word, such as instead of apricot the word peach is recalled. Authors state that, this phenomenon further reinforces the idea of being similar to human mind by showing the similarities with deep dyslexia that occurs in adults with certain brain damage.
\subsection{Language Modeling}
Language modeling is the task of predicting the next word given a sequence of words. Formally, it is the prediction of the next word's probability distribution given a sequence of words (Equation \ref{eq:language-model-1}).
\begin{equation}
\label{eq:language-model-1}
P(x_{t+1}|x_{t},...,x_{1})
\end{equation}
In an alternative interpretation, a language model assigns a probability to a sequence of words. The probability calculation can be formulated as the product of conditional probabilities in each subsequent step having the assumption that they are independent (Equation \ref{eq:language-model-2}).
\begin{equation}
\label{eq:language-model-2}
\begin{split}
    P(x_{1},...,x_{t}) &= P(x_{1})P(x_2|x_1)P(x_3|x_2,x_1)...P(x_t|x_{t-1},...,x_{1})\\
    &=\prod_{i=1}^t P(x_i|x_{i-1},...,x_{1})    
\end{split}
\end{equation}
In traditional language modeling, the next word's probability is calculated based on the statistics of n-gram occurrences. n-grams are $n$ consecutive words. In n-gram language models \cite{ngram1,ngram2}, an n-gram's probability is computed depending on the preceding $n-1$ words instead of using the product of conditional probabilities of bi-grams, tri-grams, etc. to simplify the computation.

n-gram language models have some issues. When the length of n-grams increases, their occurrence becomes sparse. This sparsity causes zero or division by zero probability values. The former one is resolved by smoothing and back-off is used to deal with the latter. Sparsity provides coarse-grained values in the resultant probability distribution. Moreover, storing all n-gram statistics becomes a major problem when the size of $n$ increases. This curse of dimensionality is a bottleneck for n-gram language models.
\subsection{Distributional Representations through Language Modeling}

\citet{Elman90} was the first to implement the distributional model proposed by \citet{Hinton1986}, in a language model. He proposes a specific recurrent neural network structure with memory, called the Elman network, to predict bits in temporal sequences. Memory is provided to network through the use of context units that are fully-connected with hidden units. 
He makes a simulation to predict bits in XOR problem. The input sequence is in the form of an input pair followed by an output bit. In the solution scheme, two hidden units are expected to represent two main patterns in the XOR truth table. That is one hidden unit should have high activation for $01$ or $10$ pattern and the other should recognize $11$ or $00$ pattern. As an alternative problem, letter sequences that are generated partially random and partially by a simple rule are tried to be learned by a recurrent neural network where hidden unit activations are used to represent word meanings.  The idea is that using such network structures, time can be modeled in an implicit way. In other words, the use of a recurrent neural network helps in learning temporal structure in language.

\citet{xu2000} create the first language model based on neural networks. Their proposed model is based on a single fully connected layer and uses one-hot vectors of words as inputs and outputs. They highlight computational cost as the major problem and in tackling the issue they mention the necessity of update mechanisms which only update those weights with non-zero input value due to one-hot encoding.

\begin{figure}[t]
    \centering
	\includegraphics[width=0.6\textwidth]{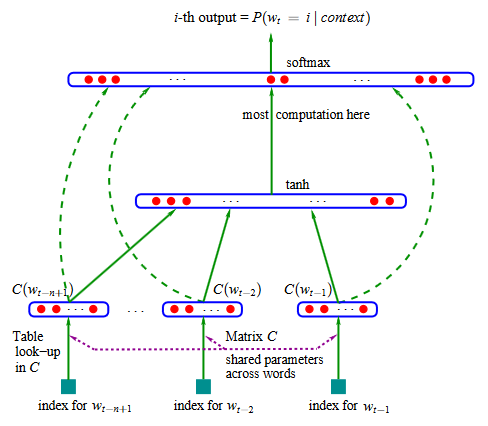}
	\caption{Neural network architecture in \citet{Bengio2003}. Taken from the original article.} 
	\label{bengio2003}
\end{figure}

\citet{Bengio2003} popularize the distributional representation idea by realizing it through a language model and lead to numerous other studies that are built on it. In their model architecture, they use a feed forward network with a single hidden layer and optional direct connections from input layer to softmax layer (Figure \ref{bengio2003}).

In addition to the advantages discussed by the aforementioned earlier works, they argue that distributional representations also break the curse of dimensionality in traditional n-gram models (\cite{ngram1}, \cite{ngram2}) where the probability of each word depends on the discrete n-grams whose numbers can exceed millions. A considerably high number of such n-grams will highly unlikely to be observed in the training set which results in sparsity problems in conditional probability calculations. A real valued feature vector representation of words will overcome this problem by working with a smooth probability function. The conditional probability of seeing a word given a context is calculated by updating the index of that word on the shared representation matrix of all the vocabulary. The probability function is smooth in that the updates that are caused by similar contexts are alike.

A second advantage of the model is the ability to capture context-based similarities. In n-gram models, the sentences "the cat is walking in the bedroom" and "a dog was running in a room" will be considered as dissimilar since they are unable to consider contexts further than $1-2$ words and have no notion of similarity among word meanings. On the other hand, in the proposed model, increasing the probability of the sentence "the cat is walking in the bedroom" will increase the probability of all the sentences below and help us generalize better:
\begin{verbatim}
"a dog was running in a room"
"the cat is running in a room"
"a dog is walking in a bedroom"
\end{verbatim}

\section{Word Embeddings with Improved Language Models}
Once it is shown that neural language models are efficiently computable by \citet{Bengio2003}, newer language models along with better word embeddings are developed successively. All of these models and their properties are summarized in Table \ref{tab:properties_early}.

\citet{alexandrescu2006} (\textbf{\textit{FNLM}}) improve the model proposed by \citet{Bengio2003} by including word-shape features such as stems, affixes, capitalization, POS class, etc. at input.

\citet{morin2005} focus on improving the performance of the earlier neural language models. Instead of using softmax and predicting the output word over the entire dictionary, they propose a hierarchical organization for vocabulary terms. A binary tree of words is created based on the IS-A relation of Wordnet hierarchy. Instead of directly predicting each word's probability, prediction is performed as a binary decision over the constructed tree's branches and leaves. This technique is an alternative to importance sampling to increase efficiency. Although the authors report exponential speed-up, the accuracy of the resultant word embeddings is a bit worse than the original method and importance sampling.

\citet{mnih2008} improve the hierarchical language model proposed by \citet{morin2005} by constructing and using a word hierarchy from distributional representations of words rather than a hierarchy built out of Wordnet. Thus, their approach is entirely unsupervised. They calculate feature vectors for words by training a hierarchical log-bilinear model (\textbf{\textit{HLBL}}) and apply EM algorithm on mixture of two Gaussians to construct a data-driven binary tree for words in the vocabulary. Authors also represent different senses of words as different leaves in the tree which is proposed in \citet{morin2005} but not implemented. Their model outperforms non-hierarchical neural models, the hierarchical neural language model that is based on Wordnet hierarchy, and the best n-gram models (\cite{ngram1}, \cite{ngram2}).

\citet{mnih2007} propose three different language models that use distributed representation of words. In Factored Restricted Boltzmann Machine (\textbf{\textit{RBM}}), they put an additional hidden layer over the distributed representation of the preceding words and exploit interactions between this hidden layer and the next word distributed representation. In temporal RBM, they further put temporal connections among hidden layer units to capture longer dependencies in the previous set of words, and finally in the log-bilinear model, called \textbf{\textit{LBL}}, they use linear dependencies between the next word and the preceding set of words. They report that the log-bilinear model outscores RBM models and also n-gram models (\cite{ngram1}, \cite{ngram2}).

\citet{Collobert2008} and \citet{almostfromscratch2011} (\textbf{\textit{C\&W}}) are among the precursors in using distributed representations in various NLP problems such as part-of-speech tagging, named entity recognition, chunking, and semantic role labeling. They propose a unified architecture for all of the problems where the words in the sentences are represented by word vectors trained from the Wikipedia Corpus in an unsupervised fashion. Although they use a feed forward architecture with a sliding window approach in word-level tasks, they utilize a convolutional neural network (CNN) architecture in semantic role labeling in order to incorporate the varying lengths of sentences, since in semantic role labeling, sliding window-based approaches don't work because target words may depend on some other far away words in a sentence. By using trained word vectors and neural network architecture, their proposed method can capture the meaning of words and succeed in various NLP tasks (almost) without using hand-crafted features. Their overall scheme is described as semi-supervised, being composed of unsupervised language modeling and other supervised tasks. 
 
\citet{mikolov2010} propose a recurrent neural network-based language model (\textbf{\textit{RNNLM}}), from where word representations can be taken. The model is able to consider contexts of arbitrary length, unlike the previous feed-forward methods where a context size should be defined beforehand. The network can learn longer dependencies. It is proved useful in tasks involving inflectional languages or languages with large vocabulary when compared to n-gram language models (\cite{ngram1}, \cite{ngram2}).

 \begin{table}[h]
	\caption{Properties of word embedding models.}\label{tab:properties_early}
	\resizebox{\columnwidth}{!}{%
	\begin{tabular}{c|c|c|c|c|c|c|c}
		\hline
		 Model & Year & Dimension & Training Corpus & NN Model & Aim & Knowledge-Base(s) & Feature(s)\\
		\hline
	    Bengio et al. \cite{Bengio2003} & 2003 & 100 & Brown & FFNN & Training & - & - \\
	    
	    Morin and Bengio \cite{morin2005} & 2005 & 100 & Brown & FFNN & Performance & Wordnet\cite{WordNet} & Hierarchical Binary Tree \\
	    
	    FNLM \cite{alexandrescu2006} & 2006 & 45-64 & \makecell{LDC ECA\cite{LDC-ECA},\\Turkish News\cite{Turkish-News}} & FFNN & Training & \makecell{LDC ECA\cite{LDC-ECA}\\Turkish News\cite{Turkish-News}} & Word Shape Features\\
	    
	    LBL \cite{mnih2007} & 2007 & 100 & APNews & RBM, FFNN & Training & - & - \\
	    
	    HLBL \cite{mnih2008} & 2008 & 100 & APNews & LBL & Performance & - & Hierarchical Binary Tree \\
	    
	    C\&W \cite{Collobert2008} & 2008 & 15-100 & Wiki & FFNN, CNN & Training & - & - \\
	    
	    RNNLM \cite{mikolov2010} & 2010 & 60-400 & Gigaword & RNN & Training & - & - \\
	    
	    CBOW \cite{mikolov2013a} & 2013 & 300-1000 & Google News & FFNN & Training & - & - \\
	    
	    Skip-Gram \cite{mikolov2013a} & 2013 & 300-1000 & Google News & FFNN & Training & - & - \\
	    
	    SGNS \cite{mikolov2013b} & 2013 & 300 & Google News & FFNN & Performance & - & Negative Sampling \\
	    
	    ivLBL/vLBL \cite{mnih2013} & 2013 & 100-600 & Wiki & LBL & Performance & - & NCE \cite{NCE} \\
	    
	    GloVe \cite{pennington2014} & 2014 & 300 & \makecell{Wiki, Gigaword,\\Commoncrawl} & LBL+coocurence Matrix & Training & - & - \\
	    
	    DEPS \cite{levy2014a} & 2014 & 300 & Wiki & CBOW & Training & \makecell{Stanford tagger\cite{stanford-tagger}\\Dependency parser\cite{dependency-parser}} & \makecell{POS,\\Dependency relation}\\
	    
	    Ling et al. \cite{ling2015a} & 2015 & 50 & Wiki & CBOW+Attn. & Training & - & - \\
	    
	    SWE \cite{liu2015b} & 2015 & 300 & Wiki & Skip-Gram & Training & Wordnet\cite{WordNet} & Ordinal Semantic Rules \\
	    
	    Faruqui et al. \cite{faruqui2015} & 2015 & - & - & - & fine-tuning & \makecell{PPDB\cite{ppdb}\\FrameNet\cite{framenet} \\WordNet\cite{WordNet}} & Semantic Relations \\
	    
	    Yin and Schütze \cite{yin2016} & 2016 & 200 & - & - & Ensemble & - & - \\
	    
	    Ngram2vec \cite{zhao2017} & 2017 & 300 & Wiki & SGNS+n-gram & Training & - & - \\
	    
	    Dict2vec \cite{tissier2017} & 2017 & 300 & Wiki & Skip-Gram & Training & \makecell{Oxford, Cambridge\\ and Collins dict.} & - \\
	
		\hline
	\end{tabular}
	}
\end{table}

\subsection{Early Word Embeddings}
Word2vec \cite{mikolov2013a} is the first neural word embedding model that efficiently computes representations to leverage the context of target words. Thus, it can be considered as the initiator of early word embeddings.
 
\citet{mikolov2013a} propose word2vec to learn high-quality word vectors. The authors removed the non-linearity in the hidden layer in the proposed model architecture of \citet{Bengio2003} to gain an advantage in computational complexity. Due to this basic change, the system can be trained using billions of words efficiently. word2vec has two variants: Continuous bag of words model (CBOW) and Skip-gram model.
 
In \textbf{\textit{CBOW}}, a middle word is predicted given its context, the set of neighboring left and right words. When the input sentence "nature is pleased with simplicity" is processed, the system predicts the middle word "pleased" given the left and right contexts. 
Every input word is in one-hot encoding where there is a vocabulary size ($V$) vector of all zeros except a one in that word's index. In the single hidden layer, the average of the neighboring left and right vectors ($w_c$) is computed to represent the context instead of applying a nonlinear transformation. As the order of words is not considered by averaging, it is named a bag-of-words model. Then the middle word's ($w_t$) probability given the context ($p(w_t|w_c)$) is calculated through softmax on the context-middle word dot product vector (Equation \ref{softmax}). Finally, the output loss is calculated based on the cross-entropy loss between the system predicted output and the ground-truth middle word.
\begin{equation}
p(w_t|w_c) = \frac{exp(w_c\cdot w_t)}{\sum\limits_{j\in V}exp(w_j\cdot w_t)}
\label{softmax}
\end{equation}

In \textbf{\textit{Skip-gram}}, system predicts the most probable context words for a given input word. In terms of a language model, while CBOW predicts an individual word's probability, Skip-gram outputs the probabilities of a set of words, defined by a given context size. Due to high dimensionality in the output layer (all vocabulary words have to be considered), Skip-gram has higher computational complexity compared to CBOW. 
Rather than traversing all vocabulary in the output layer, Skip-gram with Negative Sampling (\textbf{\textit{SGNS}}) \cite{mikolov2013b} formulates the problem as a binary classification where one class represents the current context's probability. In contrast, the other class is connected to all other vocabulary terms' occurrence probability in the present context. In the latter probability calculation, a negative sampling method is incorporated \cite{mnih2012}, which is influenced by Noise Contrastive Estimation (NCE) \cite{NCE}, to speed up the training process. As vocabulary terms are not distributed uniformly in contexts, sampling is performed from a distribution where the order of frequency of vocabulary words in corpora is taken into consideration. SGNS incorporates this sampling idea by replacing the Skip-gram's objective function. The new objective function (Equation \ref{SGNS_objective}) depends on maximizing $P(D=1|w,c)$ where $w, c$ is the word-context pair. This probability denotes the probability of $(w, c)$ coming from the corpus data. Additionally, $P(D=0|u_i,c)$ should be maximized if $(u_i,c)$ pair is not included in the corpus data. In this condition, $(u_i, c)$ pair is sampled, as the name suggests negative sampled $k$ times.\\
\begin{equation}
\sum_{w,c}\left(\log\sigma\left(\overrightarrow{w}\cdot\overrightarrow{c} \right)\right) + \sum_{i=1}^{k}\left(\log\sigma\left(\overrightarrow{-u_{i}}\cdot\overrightarrow{c} \right)\right)
\label{SGNS_objective}
\end{equation}

Both word2vec variants produced word embeddings that can capture multiple degrees of similarity including both syntactic and semantic regularities.

\citet{mnih2013} introduce speedups to the CBOW and Skip-gram models \cite{mikolov2013a}, called \textbf{\textit{vLBL}} and \textbf{\textit{ivLBL}}, by using noise-contrastive estimation (NCE) for the training of the unnormalized counterparts of these models. Training of the normalized model has a high cost due to the normalization over the whole vocabulary (the denominator term in Equation \ref{softmax}). NCE trains the unnormalized model by adapting a logistic regression classifier to discriminate between the samples under the model and the samples from a noise distribution. Thus, the computational cost and accuracy become dependent on the number of noise samples. With the relatively small number of noise samples, the same accuracy level with the normalized models is achieved in considerably shorter training times.

\citet{pennington2014} combine global matrix factorization and local context window-based prediction to form a global log bilinear model called \textbf{\textit{GloVe}}. GloVe uses ratios of co-occurrence probabilities of words as weights in its objective function to cancel out the noise from non-discriminative words. As distinct from CBOW and Skip-gram \cite{mikolov2013a}, instead of cross-entropy, GloVe uses the weighted least squares regression in its objective function. For the same corpus, vocabulary, window size, and training time, GloVe consistently outperforms word2vec.

\citet{zhao2017} (\textbf{\textit{ngram2vec}}) improve word representations by adding n-gram co-occurrence statistics to the SGNS \cite{mikolov2013b}, GloVe \cite{pennington2014}, and PPMI models \cite{levy2015}. In order to incorporate these statistics into the SGNS model, instead of just predicting the context words, they also predict the context n-gram of words. In order to add it to the other systems, they just add n-gram statistics to the co-occurrence matrix of words. They show improved scores over the models that they are built upon.

\citet{levy2014a} argue that although the word embeddings with Skip-gram are able to capture very useful representations, they also learn from unwanted co-occurrences in the context, e.g. \textit{Australian} and \textit{discovers} in the sentence \textit{"Australian scientist discovers stars with telescope"}. In order to create a different context, they use dependency trees to link each word in the sentence to the other according to the relations they have. Their experimental results show that while their model (\textbf{\textit{DEPS}}) is significantly better at representing syntactic relationships, it is worse at finding semantic relationships.
In this work, they also share a non-trivial interpretation of how word embeddings learn representations, which is very rare in neural network solutions, by examining the activations of context for specific words.

\citet{ling2015a} augment CBOW \cite{mikolov2013a} with an attention model in order to solve the shortcomings of it: Inability to account for word order and lack of treating the importance of context words differently. They show that their method can obtain better word representations than CBOW while still being faster than its complementary model Skip-gram \cite{mikolov2013a}.

\citet{yin2016} put forward the idea of ensembling the existing embeddings in order to achieve performance enhancement and improved coverage on the vocabulary. They propose four different ensemble approaches on five different word embeddings: Skip-Gram \cite{mikolov2013b}, Glove \cite{pennington2014}, Collobert\&Weston \cite{Collobert2008}, Huang \cite{huang2012}, and Turian \cite{Turian2010}. The first method CONC simply concatenates the word embeddings from five different models. SVD reduces the dimensionality of CONC. 1toN creates metaembeddings and 1to$N^+$ creates out of vocabulary (OOV) words for individual sets by randomly initializing the embeddings for OOVs and the metaembeddings,
then uses a setup similar to 1toN to update metaembeddings as well as OOV embeddings. They also propose a MUTUALLEARNING method to solve OOV problem in CONC, SVD, and 1toN. They show that the ensemble approach outperforms individual embeddings on similarity, analogy, and POS tagging tasks.

There have been some work to improve early word embeddings through knowledge bases.

\citet{liu2015b} (\textbf{\textit{SWE}}) try to improve word embeddings by subjecting them with ordinal knowledge inequality constraints. They form three different types of constraints: 
\begin{enumerate}
\item Synonym-antonym rule: A synonym of a word should be more similar than an antonym. They find these pair of words from the WordNet \cite{WordNet} synsets.
\item Semantic category rule: Similarity of words that belong to the same category should be larger than the similarity of words that are in different categories. i.e. (hacksaw, jigsaw) similarity should be greater than (hacksaw, mallet).
\item Semantic hierarchy rule: Shorter distances in hierarchy should infer larger similarities between words compared to longer distance cases. i.e (mallet, hammer) similarity should be larger than (mallet, tool).
\end{enumerate}

The last two rules are constructed from the hypernymy-hyponymy information from Wordnet.
They combine these constraints with the Skip-gram algorithm \cite{mikolov2013b} to train word embeddings and show that they can improve upon the baseline algorithm.

\citet{faruqui2015} aim to improve word embeddings with information from lexicons with a method called retrofitting. They use a word graph where each word is a vertex, and each relation in the knowledge-base is an edge between words. Their algorithm brings closer the words that are shown to be connected in the word graph and words that are found to be similar from the text. In other words, while they bring closer the words related in synsets, they also preserve the similarity in the underlying pre-trained word embeddings (Skip-gram \cite{mikolov2013a}, GloVe \cite{pennington2014}, etc.). They use various knowledge-bases such as PPDB \cite{ppdb}, WordNet \cite{WordNet}, and FrameNet \cite{framenet}.

\citet{tissier2017} (\textbf{\textit{dict2vec}}) improve word2vec \cite{mikolov2013b} by incorporating dictionary information in the form of strong and weak pair of words into the training process. If a word $a$ is in the definition of the word $b$ in dictionary and $b$ is in the definition of $a$ too, then it is a strong pair. On the other hand, if $a$ is in the definition of $b$ but $b$ is not in the definition of $a$, then they form a weak pair. The authors add this positive sampling information into the training process proportional to hyperparameters.

Despite the success of these earlier word embeddings, there were still many limitations in terms of the accuracy of representations, each of which is targeted by many research works. In the succeeding subsections, we discuss these limitations (such as morphology, senses, antonymy/synonymy, and so on) with the proposed solutions from the literature.

\subsection{Embeddings Targeting Specific Semantic Relations}
Although the initial word embedding models successfully identified semantic and syntactic similarities of words, they still need to be improved to address specific semantic relations among words such as synonymy-antonymy and hyponymy-hypernymy. 
To illustrate, consider the sentences "She took a sip of hot coffee" and "He is taking a sip of cold water." The antonyms "cold" and "hot" are deemed to be similar since their context is similar. Therefore, it becomes an issue to differentiate the synonyms "warm" and "hot" from the antonyms "cold" and "hot" considering they have similar contexts in most occurrences.

\begin{table*}[h]
	\centering
	\caption{Embeddings targeting specific semantic relations.}\label{tab:properties_sense}	
\resizebox{\columnwidth}{!}{%
	\begin{tabular}{c|c|c|c|c|c}
		\hline
		 Work & Base Model & Year & Knowledge-Base & \makecell{Morphological\\Features} & Specific Semantic Relations\\
		\hline
		
		
		
		
		
		
		dLCE \cite{nguyen2016} & SGNS \cite{mikolov2013b}) & 2016 & WordNet \cite{WordNet} and Wordnik & \xmark & Synonym-Antonym\\
		
		\citet{mrksic2016} & \makecell{GloVe \cite{pennington2014} and \\paragram-SL999 \cite{wieting2015}}& 2016 & WordNet \cite{WordNet} and PPDB 2.0 \cite{ppdb} & \xmark & Synonym-Antonym\\
		
		\citet{vulic2017} & SGNS \cite{mikolov2013b}) & 2017 & \xmark & \cmark & Synonym-Antonym\\
		
		\citet{Yu2015} & \xmark & 2015 & Probase \cite{probase} & \xmark & Hyponym-Hypernym\\
		
		\citet{tuan2016} & \xmark & 2016 & WordNet \cite{WordNet} & \xmark & Hyponym-Hypernym\\
		
		\citet{nguyenKA2017} & SGNS \cite{mikolov2013b}) & 2017 & WordNet \cite{WordNet} & \xmark & Hyponym-Hypernym\\
		
		\citet{wang2019} & Skip-gram \cite{mikolov2013a} & 2019 & \xmark & \xmark & \makecell{Synonym-Antonym,\\Hyponym-Hypernym, Meronym}\\
		\hline
	\end{tabular}
}
\label{tab:specific-semantic-relations}
\end{table*}

Table \ref{tab:specific-semantic-relations} presents the main approaches addressing synonym-antonym relations, hyponym-hypernym relations, and a study covering all types of relations.

\citet{nguyen2016} propose a weight update for SGNS \cite{mikolov2013b} to identify synonyms and antonyms from word embeddings. Their system (\textbf{\textit{dLCE}}) increases weights if there is a synonym in the context and makes a reduction in the case of an antonym. In order to come up with a list of antonyms and synonyms, they use WordNet \cite{WordNet} and Wordnik. They report state-of-the-art results in similarity tasks and synonym-antonym distinguishing datasets.

\citet{mrksic2016} propose a counter-fitting method to inject antonymy (REPEL) and synonymy (ATTRACT) constraints into vector space representations to improve word vectors. The idea behind the ATTRACT rule is that synonymous words should be closer to each other than any other word in the dictionary. Similarly, the REPEL constraint assumes that an antonym of a word should be farther away from the word than any other word in the dictionary. As knowledge-bases, they use WordNet \cite{WordNet} and PPDB 2.0 \cite{ppdb}, and as pre-trained word vectors they use GloVe \cite{pennington2014} and paragram-SL999 \cite{wieting2015}. They report state-of-the-art results on various datasets.

\citet{vulic2017} use ATTRACT and REPEL constraints on pretrained word embeddings. Their algorithm aims to pull together ATTRACT pairs while pushing REPEL pairs apart. To form the ATTRACT and REPEL constraints, the inflectional and derivational morphological rules of four languages are used; English, Italian, Russian, and German. ATTRACT constraints consist of suffixes such as (-s, -ed, -ing) to create ATTRACT word pairs such as (look, looking), (create, created).  On the other hand, REPEL constraints consist of prefixes like (il, dis, anti, mis, ir,..) to create REPEL word pairs such as (literate, illiterate), (regular, irregular).
In order to balance the changes they make to the original embeddings (they use SGNS \cite{mikolov2013b}), there is a third constraint that tries to pull word embeddings to their original positions.

In their work, \citet{Yu2015} train term embeddings for hypernymy identification. They use Probase \cite{probase} as their training data for hypernym/hyponym pairs and impose three constraints on the training process: 1) hypernyms and hyponyms should be similar to each other (dog and animal), 2) co-hyponyms should be similar (dog and cat), 3) co-hypernyms should be similar (car and auto). They create a neural network architecture to update word embeddings without optimizing parameters. They use 1-norm distance as a similarity measure. They use an SVM on the output term embeddings to decide whether a word is a hypernym/hyponym to another word.

\citet{tuan2016} aim to identify is-a relationship through neural network architecture. First, they extract hypernyms and hyponyms using the relations in WordNet \cite{WordNet} to form a training set. Second, they create (hypernym, hyponym, context word) triples by finding all sentences in the dataset containing two hypernym/hyponyms found in the first step and using the words between the hypernym and hyponym as context words. Then, they give hyponym and context words as input to the neural network and try to predict the hypernym by aggregating them with a feed-forward neural network. The resultant hypernym, hyponym pairs along with an offset vector are given to SVM to predict whether there is an is-a relationship or not. The authors state that since their method takes context words into account, their embeddings have good generalization capability and are able to identify unseen words.

\citet{nguyenKA2017} aim to learn hierarchical embeddings for hypernymy. They leverage hypernymy-hyponymy information from WordNet \cite{WordNet} and propose objective functions over/above SGNS embeddings \cite{mikolov2013b} to move hypernymy-hyponymy pairs closer. The first objective function is based on the distributional inclusion hypothesis, while the second adopts distributional informativeness. They also propose an unsupervised hypernymy measure to be used by their hierarchical embeddings. In the proposed hypernymy measure, the cosine similarity between the hypernym and hyponym vectors (to detect the hypernymy) is multiplied by the hypernym to hyponym magnitude ratio (to account for the directionality of the relation by the assumption that hypernyms are more general terms, being more frequent and thus having a large magnitude compared to hyponyms). Their evaluation also tests the generalization capability of their hypernymy solution, which proves that the model learns rather than memorizes prototypical hypernyms.

\citet{wang2019} propose a neural representation learning model for predicting different types of lexical relations, e.g., hypernymy, synonymy, meronymy, etc. Their solution avoids the "lexical memorization problem" because relation triples' embeddings are learned rather than computing those relations through individual word embeddings. In order to learn a relation embedding for a pair of words, they use the Skip-gram model \cite{mikolov2013a} over the neighborhood pairs where the similarity between pairs is defined on hyperspheres. Their lexical relation classification results verify the effectiveness of their approach.

\subsection{Sense Embeddings}

Another drawback of early word embeddings is they unite all the senses of a word into one representation. In reality, however, a word gets meaning in its use and can mean different things in varying contexts. For example, even though the words "hot" and "warm" are very similar when they are used to refer to temperature levels, they are not similar in the sentences "She took a sip of hot coffee" and "He received a warm welcome". In the transition period to contextual embeddings, different supervised and unsupervised solutions are proposed for having sense embeddings. 

\citet{schutze1998} was the first work aimed at identifying senses in texts. He defines the problem of word sense discrimination as the decomposition of a word's occurrences into same sense groups. This definition is unsupervised in its nature. When the issue becomes labeling those sense groups, the task becomes a supervised one and is named as word sense disambiguation. The reader can refer to \citet{navigli2009} for a comprehensive survey on word sense disambiguation and \citet{camacho2018} for an in-depth examination of sense embedding methods and their development.

Table \ref{tab:sense-embeddings} provides a classification of the studies that we analyze in this section. The classification dimensions include unsupervised/supervised, topical or not, knowledge base, probabilistic approach, exploiting syntactic information or not, and neural network (NN) model.

\begin{table*}[t]
	\centering
	\caption{Sense embeddings.}\label{tab:sense-embeddings}
\resizebox{\columnwidth}{!}{	
	\begin{tabular}{c|c|c|c|c|c}
    \multicolumn{6}{c}{\textbf{Unsupervised}}\\
    \hline
    \multicolumn{6}{c}{R\&M\cite{reisinger2010}}\\
    \hline
    \multicolumn{6}{c}{}\\
    \multicolumn{6}{c}{\textbf{Supervised}}\\
    
    \hline
	Work & Topical & Knowledge Base & Probabilistic & Syntactic & NN Model\\
	 &  &  &  & Information & \\
	
	\hline

    \citet{huang2012} & \xmark & \xmark & Spherical k-means & \xmark & \makecell{ Custom Language Model using \\ both local and global context } \\ 
    
    \citet{pelevina2016} & \xmark & \xmark & Graph clustering on ego network & \xmark & CBOW \cite{mikolov2013a}\\ 
    
    TWE \cite{liu2015a} & \cmark & \xmark & LDA \cite{blei2003} & \xmark & Skip-gram \cite{mikolov2013a}\\ 
    SenseEmbed \cite{iacobacci2015} & \xmark & BabelNet \cite{BabelNet} & \xmark & \xmark & CBOW \cite{mikolov2013a}\\  
    
    \citet{Chen2015} & \xmark & WordNet \cite{WordNet} & Context clustering & \xmark & CNN\\  
    
    \citet{jauhar2015} & \xmark & WordNet \cite{WordNet} & Expectation-Maximization (EM) & \xmark & Skip-gram \cite{mikolov2013a}\\ 
    
    \citet{chen2014} & \xmark & WordNet \cite{WordNet} & \xmark & \xmark & Skip-gram \cite{mikolov2013a}\\
    
    \citet{tian2014} & \xmark & \xmark & Mixture of Gaussians (EM) & \xmark & Skip-gram \cite{mikolov2013a}\\
    
    \citet{pina2015} & \xmark & SALDO \cite{borin2013} & \xmark & \xmark & Skip-gram \cite{mikolov2013a}\\
    
    MSSG \cite{neelakantan2014} & \xmark & \xmark & \cmark & \xmark & Skip-gram \cite{mikolov2013a}\\
    
    SAMS \cite{cheng2015} & \xmark & \xmark & \xmark & \cmark & Recursive Neural Network\\
    
    \citet{li2015} & \xmark & \xmark & Chinese Restaurant Process (CRP) & \xmark & CBOW-Skip-gram \cite{mikolov2013a}, SENNA \cite{almostfromscratch2011}\\ 
    
    MSWE \cite{nguyenDQ2017} & \cmark & \xmark & LDA \cite{blei2003} & \xmark & Skip-gram \cite{mikolov2013a}\\
    
    \citet{guo2014} & \xmark & \xmark & Affinity Propagation Algorithm & \xmark & RNNLM model \cite{mikolov2010}\\
    
    LSTMEmbed \cite{iacobacci2019} & \xmark & BabelNet \cite{BabelNet} & \xmark & \xmark & LSTM\\
    
    \citet{kumar2019} & \xmark & \makecell{Knowledge Graph\\Embedding} & \xmark & \xmark & \makecell{ Framework consisting of \\ different types of Encoders } \\
    
		\hline
	\end{tabular}
}
\end{table*}

At the outset, unsupervised learning is used to discriminate the different senses of a word.

\citet{reisinger2010} propose a multi-prototype based word sense discovery approach. In their approach (\textbf{\textit{R\&M}}), a word's all occurrences are collected as a set of feature vectors and are clustered by a centroid-based clustering algorithm. The resultant clusters (fixed number) for each word are expected to capture meaningful variation in word usage rather than matching to traditional word senses. They define the similarity of words $A$ and $B$ as the "maximum cosine similarity between one of A's vectors and one of B's vectors" and provide experimental evidence on similarity judgments and near-synonym prediction. Moreover, variance in the prototype similarities is found to predict variation in human ratings.

Following \citet{reisinger2010}, \citet{huang2012} also aim at creating multi-prototype word embeddings. They compute vectors using a feed forward neural network architecture with one layer to produce single prototype word vectors and then perform spherical k-means to cluster them into multiple prototypes. They also introduce the idea of using global context where the vectors of words in a document are averaged to create a global semantic vector. The final score of embeddings is then calculated as the sum of scores of each word vector along with the global semantic vector.

The authors also argue that available test sets for similarity measurements are not sufficient for testing multi-prototype word embeddings because the scores of word pairs in those test sets are given in isolation, which lacks the contextual information for senses. Therefore, they introduce a new test set in which the word pairs are scored within a context by mechanical turkers, where context is usually a paragraph from Wikipedia that contains the given word. Finally, they show that their model is capable of outperforming the former models when such a test set is used, although its performance is similar to others in previous test sets.

\citet{pelevina2016} aim at creating sense embeddings without using knowledge bases. Their model takes the existing single-prototype word embeddings and transforms them into multi-prototype sense embeddings by constructing an ego network and performing graph clustering over it. In fact, the senses of a word they learn do not have to correspond to the senses of that word in the dictionary. They evaluate their method on their crowd-sourced dataset.

\citet{liu2015a} propose three different methods to create topical embeddings (\textbf{\textit{TWE}}). They create their topical embeddings without the use of any knowledge base, but instead rely on LDA \cite{blei2003} to find the topics of each document the word occurs in. Topical embeddings they create are similar to sense embeddings with the only difference being that the number of topics may not correspond to the number of senses in the dictionary. 

In their first model, named TWE-1, they learn word embeddings and topic embeddings separately and simultaneously with the skip-gram method by treating topic embeddings as pseudo-words, which appear in all the positions of words under this topic. The sense embeddings of a word $w$ for topic $t$ are then constructed by concatenating the word embedding $w$ with the corresponding topic embedding $t$. Their second model TWE-2 treats word embeddings and topic embeddings as tuples and train them together. This method may lead to sparsity issues since some words on a specific topic may not be frequent. The last method they propose, TWE-3, also train word and topic embeddings together, but this time the weights of embeddings are shared over all word-topic pairs. They show that the TWE-1 method gives the best results overall, and the independence assumption between words and topics in the first model is given as the reason behind its performance.

Exploiting vast information in knowledge bases to learn sense representations has proved useful. The approaches that rely mainly on knowledge bases to compute sense embeddings include \citet{iacobacci2015}, \citet{Chen2015}, \citet{jauhar2015}, and \citet{chen2014}.

\citet{iacobacci2015} (\textbf{\textit{SenseEmbed}}) use BabelNet \cite{BabelNet} as a knowledge-base to retrieve word senses and to tag words with the correct sense. They train the sense-tagged corpora on the CBOW architecture and achieve state-of-the art results in various word similarity and relatedness datasets.

\citet{Chen2015} also use a knowledge-base (WordNet) to solve the sense-embedding problem. They use CNN to initialize sense-embeddings from the example sentences of synsets in WordNet. Then, they apply context clustering to create distributed representations of senses. The representations they obtain achieve promising results.

\citet{jauhar2015} propose two models for learning sense-embeddings using ontological resources like WordNet \cite{WordNet}. In their first model, they retrofit pretrained embeddings by imposing two conditions on them: pulling together the words that are ontologically-related (by using the graphs constructed from the relationships in WordNet) and leveraging the tension between sense-agnostic neighbors from the same graph. They implement the first method over Skip-gram \cite{mikolov2013b} and \citet{huang2012} and show that their method can improve the success of the previous methods. Their second method constructs embeddings from scratch by training them with an Expectation-Maximization (EM) objective function that pulls together ontologically-related words similar to the first model and finds the correct sense of the word from WordNet and creates a vector for each sense. 

\citet{chen2014} propose a unified model for word sense representation (WSR) and word sense disambiguation (WSD). The main idea behind this is that both models may benefit from each other. Their solution is composed of three steps: First, they initialize single-prototype word vectors using Skip-gram \cite{mikolov2013b} and initialize the sense embeddings using the glosses in WordNet \cite{WordNet}. They take the average of words in WordNet synset glosses to initialize the sense embeddings. Second, they perform word sense disambiguation using some rules on the given word vectors and sense vectors. Finally, using the disambiguated senses, they learn sense vectors by modifying the Skip-gram objective such that both context words and context words' senses must be optimized given the middle word in context.

\citet{tian2014} propose a probabilistic approach to provide a solution to sense embeddings. They improve the Skip-gram algorithm by introducing the mixture of Gaussians idea to represent the given middle word in context in the objective function. Every Gaussian represents a specific sense, and the mixture is their multi-prototype vector. The number of Gaussians, in other words, the number of senses, is a hyperparameter of the model. They use Expectation-Maximization (EM) algorithm to solve the probabilistic model.

\citet{pina2015} extend the Skip-gram \cite{mikolov2013a} method to find sense representations of words. They get the number of senses from a knowledge-base and for each word in the training corpus, they find the most probable sense by using the likelihoods of context words. They only train the sense with the highest probability. They train their system on Swedish text and measure their success by comparing the senses to the ground-truth in the knowledge-base (SALDO \cite{borin2013}).

\citet{neelakantan2014} (\textbf{\textit{MSSG}}) also aim at creating word vectors for each sense of a word. Unlike most other models, they do it by introducing the sense prediction into the neural network and jointly performing sense vector calculation and word sense discrimination. Their first model relies on Skip-gram and induces senses by clustering the context word representations around each word. Then, the word is assigned to the closest sense by calculating the distance to the sense clusters' centers. Here the count of clusters is the same for all words and is a hyperparameter. Their second model is a non-parametric variant of the first one where a varying number of senses is learned for each word. A new cluster (sense) for a word type is created with probability proportional to the distance of its context to the nearest cluster (sense). They show that their second method can outperform the first since it can better learn the senses' nature.

\citet{cheng2015} consider capturing syntactical information to better address senses. They use recursive neural networks on parsed sentences to learn sense embeddings. Each input is disambiguated to its sense by calculating the average distance of the words' embeddings in the sentence to sense cluster means. They define two negative sampling methods to train the network. One negative example is created to swap the target word with a random word (as in \cite{mikolov2013b} and \cite{NCE}), another negative sampling changes the order of words in a sentence, which further enforces the model (\textbf{\textit{SAMS}}) to learn syntactic dependencies.

\citet{li2015} decide the number of senses in an unsupervised fashion by using the Chinese Restaurant Process (CRP). They combine CRP with neural network training methods by determining the sense of a word by looking at its context. They also compare sense-embedding methods with single-prototype models across various NLP tasks to see if they are beneficial. They state that in some tasks (POS tagging, semantic relatedness, semantic relation identification), sense-embeddings outperform single-prototype methods. Still, they fail to improve their scores on some other tasks (NER, sentiment analysis).

Instead of getting the number of senses from a knowledge-base, \citet{nguyenDQ2017} (\textbf{\textit{MSWE}}) use LDA \cite{blei2003} to find word to topic and topic to document probability distributions. Here the number of topics is a parameter to the model. They train different weights for each sense of a word using two different optimization methods. The first model learns word vectors based on the most suitable topic. On the other hand, their second model considers all topics to learn them. They conclude that this second method can be considered as a generalization of the Skip-gram model \cite{mikolov2013a} given the fact that it behaves as Skip-gram if the mixture weights are set to zero.

\citet{guo2014} exploit bilingual resources to find sense embeddings, motivated by the idea that if a word in a source language translates into multiple words in a target language, that means different words in the target language corresponds to a sense in the source language. For this purpose, they use Chinese to English translation data to induce senses in an unsupervised fashion. They represent the initial words with the word embeddings from C\&W \cite{Collobert2008} and use the affinity propagation algorithm to cluster the translated words into dynamic clusters, which means that their method can learn a different number of senses for each word. Then, they use the RNNLM model \cite{mikolov2010} to train the sense embeddings.

\citet{iacobacci2019} propose an LSTM-based architecture (\textbf{\textit{LSTMEmbed}}) to jointly learn word and sense embeddings. Input contexts are provided from semantically annotated data, and one bidirectional LSTM processes the left context while another one handles the right one. As an extra layer, the concatenation of both outputs is linearly projected into a dense representation. Then, the optimization objective tries to maximize the similarity between the produced dense output and pretrained word embeddings from SGNS. Consideration of these pretrained word embeddings in the final phase increases the vocabulary use of the proposed system. Their experiments on the word to sense similarity and word-based semantic evaluations prove the usefulness of their approach.

\citet{kumar2019} propose a framework that combines a context encoder with a definition encoder to provide sense predictions for out of vocabulary words. In the case of rare and unseen words, most word sense disambiguation (WSD) systems rely on the most frequent sense (MFS) on the training set.  In the part of the definition encoder, sentence encoders along with knowledge graph embeddings are utilized. Here instead of using discrete labels for senses, the score for each sense in the inventory is calculated by the dot product of the sense embedding with the projected context-aware embedding.

\subsection{Morpheme Embeddings}
The quest for morphological representations is a result of two important limitations of earlier word embedding models. The first point is, words are not the smallest units of meaning in languages, morphemes are. Even if a model does not see the word \textit{unpleasant} in the training it should be able to deduce that it is the negative form of \textit{pleasant}. Word embedding methods that don't take morphological information into account can not produce any results in such a situation. The second limitation is the data scarcity problem of morphologically rich languages and agglutinative languages. Unlike English, morphologically rich languages have many more noun and/or verb forms inflected by gender, case, or number, which may not exist in the training corpora. The same thing is also valid for agglutinative languages in which words can have many forms according to the suffix(es) they take. Therefore, models that take morphemes/lexemes into account is needed.

\begin{table}[t]
	\centering
	\caption{Morpheme embedding models.}\label{tab:morphemeEmbeddings}
	\resizebox{\columnwidth}{!}{
	\begin{tabular}{c|c|c|c|c|c}
		\hline
		 Model & Year & Training Corpus & Knowledge-Base & NN Model & Dimension\\
		\hline
		
		Luong et al. \cite{luong2013} & 2013 & Wiki & Morfessor\cite{morfessor} & recNN & 50 \\
		
		CLBL \cite{botha2014} & 2014 & ACL MT & Morfessor\cite{morfessor} & LBL & - \\
		
		Qiu et al. \cite{qiu2014} & 2014 & Wiki & Morfessor\cite{morfessor},Root,Syllable\cite{syllable} & CBOW & 200 \\
		
		Bian et al. \cite{bian2014} & 2014 & Wiki & \makecell{Morfessor\cite{morfessor}, WordNet\cite{WordNet},\\Freebase\cite{freebase}, Longman Dict.} & CBOW & 600\\
		
		CharWNN \cite{santos2014} & 2014 & Wiki & - & CNN & 100\\ 
		
		KNET \cite{cui2015} & 2015 & Wiki & Morfessor\cite{morfessor}, Syllable\cite{syllable} & Skip-Gram & 100\\
		
		AutoExtend \cite{rothe2015} & 2015 & Google News & WordNet \cite{WordNet} & Autoencoder & 300 \\
		
		Morph-LBL \cite{cotterell2015} & 2015 & TIGER \cite{tiger} & TIGER \cite{tiger} & LBL & 200\\
		
		Soricut and Och \cite{soricut2015} & 2015 & Wiki & - & Skip-Gram & 500\\
		
		C2W \cite{ling2015b} & 2015 & Wiki & - & biLSTM\cite{biLSTM} & 50\\
		
		Cotterell et al. \cite{cotterell2016} & 2016 & Wiki & CELEX \cite{celex} & GGM & 100 \\
		
		Fasttext \cite{bojanowski2016} & 2016 & Wiki & - & Skip-Gram & 300\\
		
		char2vec \cite{cao2016} & 2016 & text8 (wiki) & - & LSTM\cite{lstm}+Attn & 256 \\
		
		\citet{kim2016} & 2016 & ACL MT & - & CNN+LSTM & 300-650\\
		
		LMM \cite{xu2018} & 2018 & Gigaword & Morfessor\cite{morfessor} & CBOW & 200 \\
		
		\hline
	\end{tabular}
	}
\end{table}

Researchers propose several ways to target morphological information in order to obtain sub-word information for solving the rare/unknown word problem of earlier word embedding methods and also to have better representations of words for morphologically rich languages. While some of the works are proposed to train embeddings directly from morphemes/lexemes, others adjust the representations of other word embedding models. Summary of these models and their properties can be seen in Table \ref{tab:morphemeEmbeddings}.
\vspace{3mm}

\subsubsection{Training Morphological Embeddings from Scratch}
There are two main ways for training morpheme embeddings from scratch: While some methods (\cite{luong2013}, \cite{botha2014}, \cite{qiu2014}, \cite{bian2014}, \cite{cui2015}, \cite{cotterell2015}, \cite{xu2018}, \cite{soricut2015}) propose to use tools or special rules for dissecting a text to its morphemes, others (\cite{bojanowski2016}, \cite{cao2016}, \cite{ling2015b}, \cite{santos2014}) prefer using characters or character n-grams as input to learn morphemes along with their representations.

\citet{luong2013}'s work is the first work that attempts to incorporate morphological information in word embeddings. They train morphological embeddings with recursive neural networks. They divide words into (prefix, stem, affix) tuples by using morfessor \cite{morfessor} and feed them to a recursive neural network. Word embeddings are then constructed by a word-based neural language model (NLM). Instead of initializing the vectors with random numbers, they initialize them with the pre-trained word embeddings from \citet{almostfromscratch2011} and \citet{huang2012} in order to focus on learning the morphemic semantics.

Similar to \citet{luong2013}, \citet{botha2014} (\textbf{\textit{CLBL}}) also use morfessor \cite{morfessor} to find the morphemes of words in text and train both the target word and context words by first factoring them into their morphemes. They learn the morphology-based word representations with an additive-LBL of their factor embeddings, e.g., surface form, stem, affixes, etc.

\citet{qiu2014} incorporate morphemes into the CBOW \cite{mikolov2013a} architecture: Instead of predicting a word from the context words, they propose to use both morphemes and words as input and for prediction. They control the relative contributions of words and morphemes with two parameters that weigh the information to be extracted from each input. They use three different tools for extracting morphemes from corpus: \textit{Morfessor} \cite{morfessor}, \textit{root}, and \textit{syllable} \cite{syllable}.

\citet{bian2014} investigate three different methods for finding better representations for words and morphemes: First, they transform CBOW \cite{mikolov2013a} into a new basis by using morphemes (segmented by using morfessor \cite{morfessor}) instead of words. They later represent words as the aggregate of the morphemes they are composed of. Second, they provide additional information to their first model by feeding semantic and syntactic information vectors as inputs along with the morpheme vectors. As semantic and syntactic information, they use synsets, syllables, syntactical transformation, and antonym and synonyms from Freebase \cite{freebase}, WordNet \cite{WordNet}, and Longman dictionaries\footnote{www.longmandictionariesonline.com}. Finally, they use syntactic knowledge (POS tagging vector) and semantic knowledge (entity vector and relation matrix) as auxiliary tasks, where they use syntactic/semantic information as outputs around the center word to be predicted. Their relation matrix consists of relations such as \textit{belong-to} and \textit{is-a} relation. They examine the effects of both semantic and syntactic information compared to the baseline model (CBOW) and report the relative effects in various tasks.

\citet{soricut2015} aim at improving word vectors and solving the rare word problem by using morphology induction. In their method, they first extract candidate morphological rules. In this step, they find the word pairs $(w1, w2)$ such that $w2$ is formed by substituting prefixes and suffixes up to $6$ characters from $w1$ (i.e., $(bored, boring)$ is produced from the rule $(suffix:ed:ing)$). Later they form their rules from word pairs. After training their embeddings with the Skip-gram method \cite{mikolov2013a}, they keep the rule if the word pair $(w1,w2)$ is similar in embedding space; otherwise, the rule is removed from the candidate rule list. Thus, they use their morphological rules to obtain representations for rare words that may or may not be in the training set.

\citet{cui2015} (\textbf{\textit{KNET}}) use co-occurrence statistics to construct word embeddings with sub-word information. They leverage four different morphological information inspired by the advances in cognitive psychology: i) edit distance similarity ii) longest common sub-string similarity, iii) morpheme similarity (share roots, affixes, etc. by using morfessor \cite{morfessor}), and iv) syllable similarity (by using hyphenation tool \cite{liang1983}). They combine the aforementioned morphological information into a relation matrix and construct morphological embeddings from it. On the other hand, they also create word embeddings by using the Skip-gram method \cite{mikolov2013b}. A combination of these two embeddings with weighted averaging is used in order to obtain the final word embeddings. Unlike most other word embedding methods, authors do not change the digits in the text with zeros; instead, they change the digits with their text counterparts to reflect the information better.

Different from other morphology-based models, \citet{cotterell2015} implement a semi-supervised approach (\textbf{\textit{MorphLBL}}) where a partially morphologically tagged dataset (TIGER dataset of German newspaper \cite{tiger}) is used. They augment the LBL model \cite{mnih2007} to both predict word and morpheme together. They also introduce a new metric for measuring the success of morphological models called MorphDist.

\citet{santos2014}, \citet{ling2015b}, \citet{bojanowski2016}, and \citet{cao2016} come up with character-based solutions instead of using a tool/knowledge-base to find morphemes in sentences.

In their work (\textbf{\textit{CharWNN}}), \citet{santos2014} use word embeddings together with character embeddings to compensate for the need for hand-crafted features in part-of-speech (POS) tagging, where the morphological structure of words plays a significant role. In their architecture, they use Skip-gram \cite{mikolov2013a} for word embeddings and train their character embeddings from scratch.

The compositional model of \citet{ling2015b}, called \textbf{\textit{C2W}}, takes the characters of a word as input and uses bidirectional-LSTM to construct word vectors by concatenating the last state of LSTM in each direction.

\citet{bojanowski2016} propose a model, called \textbf{\textit{Fasttext}}, that takes character $3$- to $6$-grams of words and represents the words with a bag of n-grams. i.e., for the word \enquote{where} the $3$-grams are: (<wh, whe, her, ere, re>), where $<$ and $>$ are special characters for denoting the beginning and end of the word, respectively. N-grams are then summed to produce word embeddings. Thus, as the model shares representations across words, it can have better representations for rare words. They perform extensive tests on morphologically rich languages to see how their model works and learns the subword information.

\citet{cao2016} aim at solving unsupervised morphology induction and learning word embeddings jointly by using bidirectional LSTMs with the Bahdanau attention \cite{bahdanau2014NMT} on characters. The output of the attention layer is fed to Skip-gram \cite{mikolov2013a} algorithm to compute word representations. They prove that the attention layer learns to split the words into multiple morphemes by showing that their algorithm outperforms other morpheme induction methods. However, it is not only designed for solving that problem. They also show that since their method (\textbf{\textit{char2vec}}) focused on finding morpheme representations through characters, it is better at tasks that measure syntactic similarity. On the other hand, they argue that their method is worse at tasks that measure semantic similarity since characters do not convey any semantic information of words alone.

To address both syntactic and semantic features, \citet{kim2016} use a mixture of character and word-level features. In their model, at the lowest level of the hierarchy, character-level features are processed by a CNN; after transferring these features over a highway network, high-level features are learned using an LSTM. Thus, the resulting embeddings show good syntactic and semantic patterns. For instance, the closest words to the word \textit{richard} are returned as \textit{eduard, gerard, edward, and carl}, where all of them are person names and have a high syntactic similarity to the query word. Due to character-aware processing, their models are able to produce good representations for out-of-vocabulary words.

\citet{xu2018} (\textbf{\textit{LMM}}) also aim at enhancing word representations with morphological information. In incorporating morphological information, the authors suggest using the latent meaning of morphemes instead of morphemes themselves. They state that although the words $incredible$ and $unbelievable$ have similar semantics, the methods based on morphemes cannot catch it. Instead, they use the latent meaning of morphemes that they extract from knowledge-bases (i.e. in=not, un=not, ible=able, able=able, cred=believe, believ=believe). They use CBOW \cite{mikolov2013a} as pretrained word embeddings and show improvements using their method on them.

\subsubsection{Adjusting the Existing Embeddings}
Among the models that adjust the pre-trained word embeddings, \citet{rothe2015} take any word embeddings and transform them into embeddings for lexemes and synsets. To do that, they use WordNet \cite{WordNet} synsets and lexemes although they note that their model (\textbf{\textit{AutoExtend}}) can get the information from the other knowledge bases such as Freebase \cite{freebase}. They consider words and synsets as the sum of their respective lexemes and enforce three constraints on the system i) synset constraint, ii) lexeme constraint, and iii) WordNet constraint (because some synsets contain only a single word). They use an autoencoder where the result of the encoding corresponds to synset vectors, and the hidden layer in encoding and its counterpart in decoding correspond to lexeme vectors. Two lexeme vectors are then averaged to produce the final lexeme embeddings. 

On the other hand, \citet{cotterell2016} use a Gaussian graphical model where word embeddings are represented as the sum of their morphemes. Their system takes the output of the other word embedding methods as input and converts them by learning their morpheme embeddings and calculating the word embeddings by summing them. They also note that it is also possible to extrapolate the embeddings of OOV words with their method since one can compute their morpheme embeddings from the same morpheme in other words.

\section{Contextual Representations}
As it is shown in the last section, many methods have been proposed for solving the deficiencies of embedding methods. Each of them is specialized on a single problem such as sense representation, morpheme representation, etc., while none of them was able to combine different aspects together into a single model, a single solution. It is the idea of \textit{contextual representations} to provide a solution that covers each aspect successfully. The main idea behind contextual representations is that words should not have a single representation to be used in every context. Instead, a representation should be calculated separately for different contexts. Contextual representation methods calculate the embedding of a word from the surrounding words each time the word is seen, contrary to the earlier methods where each word is represented with a fixed vector of weights. This leads to an implicit solution to many problems such as sense representations, antonymy/synonymy, and hypernymy/hyponymy, since now multi-sense words can have different representations according to their context. Furthermore, it has also been proposed to use characters as input which also incorporates the sub-word information into embeddings. Therefore, contextual representation models, described below, are able to incorporate different aspects together into a single model. \citet{contextualsurvey} examine contextual embeddings in detail by comparing their pre-training methods, objectives, and downstream learning approaches.

In such a first attempt to create contextual representations, \citet{melamud2016} developed a neural network architecture based on bidirectional-LSTMs to learn context embeddings with target word embeddings jointly. They feed words to a $2$-layer bidirectional LSTM network in order to predict a target word in a sentence. They use sentences as context and feed the left side of the target word to left to right (forward) biLSTM and feed the right side of the target word to right to left (backward) biLSTM. To jointly learn context and target word embeddings, they use the Skip-gram objective function sampled on context-word occurrences. Furthermore, they show that this is equivalent to the factorization of a context-target word co-occurrence matrix. Although the previous word embedding models create both context and target word embeddings, they only use target-target similarity as representations and ignore the context embeddings. In this work, the authors also use context-context and context-target to show that contextual embeddings can significantly improve NLP systems' performance. They also show that since bidirectional LSTM structures can learn long-term contextual dependencies, their model, context2vec, is able to differentiate polysemous words with a high success rate.

CoVe \cite{McCann2017} uses Glove \cite{pennington2014} as the initial word embeddings and feeds them to a machine translation architecture to learn contextual representations. The authors argue that pre-training the contextual representations on machine learning tasks, where there are vast amounts of data, can lead to better contextual representations to transfer learning to other downstream tasks. They concatenate the output of the encoder of a machine translation model (as contextual embeddings) with the GloVe embeddings to construct their final word representations.

Using language modeling and learning word representations as to the pre-training objective, then fine-tuning the architecture to downstream tasks is first proposed by \citet{Dai2015} and \citet{ulmfit}. While \citet{Dai2015} propose to use RNNs and autoencoders to tackle the issue, ULMFiT \cite{ulmfit} introduces novel fine-tuning ideas such as discriminative fine-tuning, slanted triangular
learning rates, and gradual unfreezing to their LSTM model, inspired from the advances in transfer learning in computer vision. After the success shown by these models, the aim is shifted from creating word representations to using their system as pre-trained models and then fine-tuning a classifier on top to perform downstream tasks.

ELMO \cite{peters2018} improves on the character-aware neural language model by \citet{kim2016}.  The architecture takes characters as input to a CNN network from where it is fed to a $2$-layer bidirectional-LSTM network to predict a target word. They show that this architecture can learn various aspects of words such as semantic, syntactic, and sub-word information. First, they show that, since the model takes characters as inputs, it is able to learn sub-word information even for the unseen words. Second, they show that while the first layer of biLSTM better captures the syntactic similarity of words, the second layer better captures the semantics. Therefore, they propose to use the different layers of the model to create word representations. They also propose to use a weighted averaging method for combining the different layers. They show that including ELMO representations can improve many state-of-the-art models in various NLP tasks.

Instead of using words as input, Flair \cite{akbik2018} uses a character-level language model to learn contextual word representations. Different from ELMO \cite{peters2018} where character level inputs are later converted into word features, in this work, the authors propose to use characters only. They feed the characters of an input string to a single layer LSTM network and predict the next character. They later form the word representation by concatenating the backward LSTM output from the beginning of the word with the forward LSTM output from the end of the word. They also try concatenating other pre-trained word vectors with their contextual representations in downstream tasks and show that this can improve the results.

BERT \cite{devlin2019} uses bidirectional transformer \cite{vaswani2017} architecture to learn contextual word representations. Different from the earlier approaches (ELMO \cite{peters2018}, \citet{melamud2016}) BERT is bidirectional. Although ELMO also considers both sides of a target word, it considers them separately as the left and right sides. Instead, BERT spans the entire sentence with both right to left and left to right transformers. To do so, without also spanning the target word, they mask the target word. Therefore, they call this model a masked language model (MLM).

In addition to the token (word) embeddings, they also use segment (sentence) embeddings and position embeddings (words' position in segments) as input which enables BERT to consider multiple sentences as context and to represent inter-sentence relations. Giving multiple sentences as input helps BERT be integrated into most downstream tasks that require inter-sentence connections such as question answering (QA) and natural language inference (NLI) easily without requiring any other architecture. For further details, the reader can refer to the work of \citet{Bertology}, which provides an in-depth survey on how exactly BERT works and what kind of information it captures during training and fine-tuning.

XLNet \cite{xlnet} is an autoregressive method that combines the advantages of two language modeling methods: Autoregressive models (i.e. transformer-XL \cite{transformerXL}) and autoencoder models (i.e. BERT \cite{devlin2019}). Specifically, it considers both sides of the target word by employing a permutation language modeling object without masking any words like BERT, which allows their model to capture also the relation between the masked word and the context words, unlike BERT.

ALBERT \cite{albert} aims at lowering the memory consumption and training times of BERT \cite{devlin2019}. To accomplish this, they perform two changes on the original BERT model: They factorize the embeddings into two matrices to use smaller dimensions, and they apply weight sharing to decrease the number of parameters. They state that weight sharing also allows the model to generalize better. They show that although they can obtain state-of-the-art results over BERT with fewer parameters, ALBERT requires a longer time to train than BERT.

RoBERTa \cite{roberta} revises the pre-training design choices of BERT \cite{devlin2019} by trying alternatives in a controlled way. Specifically, dynamic masking for the Masked Language Model (MLM), input format of entire sentences from a single document with the Next Sentence Prediction (NSP) loss removed, and byte-level Byte Pair Encoding (BPE) vocabulary give better performance. Moreover, they extend the training set size and the size of mini-batches in training. As a result, RoBERTa \cite{roberta} achieves state-of-the-art results in GLUE, RACE, and SQuAD benchmarks.

In their work, called ERNIE, \citet{ernie} improve on BERT by introducing two knowledge masking strategies into their masked language modeling. In addition to masking out random words in the sentence, they mask phrases and named entities to incorporate real-world knowledge into language modeling/representation. In their successive work, ERNIE 2.0 \cite{ernie2}, they implement continual multi-task learning. Including the one in ERNIE, they define seven pre-training tasks categorized into word-aware, structure-aware, and semantic-aware pre-training tasks, aiming to capture lexical, syntactic, and semantic relations, respectively.

GPT and its variants rely on a meta-learner idea using a conditional language model in diverse NLP tasks. This conditional language model predicts the next word conditioned both on an unsupervised pre-trained language model and the previous set of words in context. In GPT-3, \citet{gpt-3} pre-train a $175$ billion parameter transformer-based language model on a sufficiently large and diverse corpus and tests its performance in zero-shot, one-shot, and few-shot settings. Their learning curves for these three settings show that a larger model better learns a task from contextual information. Authors apply task-specific input transformations, e.g., delimiting context and question from the answer in reading comprehension, to test the model's performance in different NLP tasks. Their few-shot results prove the effectiveness of their approach by outperforming state-of-the-art on LAMBADA language modeling dataset \cite{lambada}, TriviaQA closed book open domain question answering dataset \cite{triviaQA}, and PhysicalQA (PIQA) common sense reasoning dataset \cite{piqa}.

\section{Performance of Word Representations}
Due to the popularity of the field, many datasets have been proposed and tested upon. In this section, we report the structure of the datasets and the performance of the aforementioned word embedding models on them.

\subsection{Datasets}
Depending on their aim, the datasets produced to measure the success of embedding models can be divided into four categories: Similarity tasks, analogy task, synonym selection tasks, and downstream tasks.

\subsubsection{Similarity Tasks} These datasets provide pairs of words whose similarity is rated by human judgments. They all use Spearman's rank correlation ($\rho$) with average human judgment to measure the performance and quality of embeddings. 
\begin{itemize}
    \item WordSim-353 (WS-353): \citet{WordSim353} produced a corpus that contains human judgements, rated from $1$ to $10$, on $353$ pairs of words.\\
    
    \item SCWS: \citet{huang2012} introduced this dataset in which the word pairs are scored by mechanical turkers within a context, which is usually a paragraph from Wikipedia that contains the given word. The reason for introducing such a dataset is that the available test sets for similarity measurements are not sufficient for testing multi-prototype word embeddings because the scores of word pairs in those test sets are given in isolation, which lacks the contextual information for senses.\\
    
    \item RG-65: This dataset, developed by \citet{RG65}, is composed of $65$ noun pairs whose similarity is rated by humans.\\
    
    \item MC-30: The dataset \cite{MC30} contains $30$ pairs of words.\\
    
    \item MEN: It \cite{MEN} contains $3000$ pairs of words together with human assigned similarity scores obtained from Amazon Mechanical Turk.\\
    
    \item YP-130: Similar to the previous test sets, YP-130 \, cite{YP130} also contains human assigned similarity scores to $130$ word pairs.\\
    
    \item RW: Unlike the previous word similarity datasets, RW \cite{luong2013} consists of $2034$ pairs of rare words which are not frequently seen in texts. The motivation behind this dataset is to provide a sufficient number of complex and rare words to test the expressiveness of morphological models since the previous datasets mostly contain frequent words that are insufficient for such tests.\\
    
    \item Simlex-999: Simlex-999 dataset \cite{simlex-999} contains $999$ pairs of words whose similarity is annotated by mechanical turkers.\\
\end{itemize}

\subsubsection{Analogy Task}
Semantic-syntactic word relationship test set (Google Analogy Task) introduced by \citet{mikolov2013a} consists of pairs of words in the form of $a$ is to $a^*$ as $b$ is to $b^*$ (such as Paris is to France as London is to England). The aim is to find $b^*$, given $a$, $a^*$, and $b$ (cosine distance is used as a distance metric to find the missing word). There are $8869$ semantic and $10675$ syntactic questions in the dataset, and the success is measured by accuracy.\\

\subsubsection{Synonym Selection Tasks}
Given a word as input, this task aims to select the most synonym-like word among the list of candidates. Accuracy (\%) is used to measure the performance.
\begin{itemize}
    \item ESL-50: Contains $50$ synonym selection questions from ESL (English as a second language) tests.\\
    
    \item TOEFL-80: Contains $80$ synonym selection questions from TOEFL (Test of English as Foreign Language) tests.\\
    
    \item RD-300: Contains $300$ synonym selection problems from Reader's Digest Power Game.\\
    
\end{itemize}

\subsubsection{Downstream Tasks}
As representations and models get better and the difference between word embedding methods and language models gets closer, experiments are shifted from similarity tasks to downstream tasks.

GLUE benchmark dataset \cite{GLUE} is introduced to provide a stable testing environment for researchers. It consists of several downstream tasks:
\begin{itemize}
    \item CoLA: The Corpus of Linguistic Acceptability \cite{CoLA} is a sentence classification task where the aim is to determine whether a sentence is linguistically acceptable or not. It contains $9594$ sentences from linguistic publications and the success is measured by Matthew' Correlation Coefficient (MCC).
    \item SST-2: The Stanford Sentiment Treebank \cite{SST-2} consists of $68.8k$ sentences from movie reviews. The aim is to classify the sentiment of sentences. Accuracy is used to measure the performance.
    \item MRPC: Microsoft Research Paraphrase Corpus \cite{MRPC} contains $5800$ pairs of sentences from news sources on the web. Each pair is annotated by humans indicating whether they are semantically equivalent or not. Performance is measured by accuracy.
    \item STS-B: Semantic Textual Similarity Benchmark  \cite{STS-B} is composed of $8628$ pairs of sentences from various sources, annotated between $1$ and $5$, determining how similar they are. Success is measured by  Spearman's rank correlation ($\rho$).
    \item QQP: Quora Question Pairs \cite{QQP} dataset contains over $400k$ question pairs where the aim is to determine whether the questions are semantically similar or not. Success is measured by accuracy. 
    \item MNLI: Multi-Genre Natural Language Inference \cite{MNLI} dataset is composed of $430k$ crowd-sourced sentence pairs annotated with entailment information. The aim is to predict whether a second sentence is a contradiction, entailment, or neutral to the first one. Accuracy is used to measure the performance.
    \item QNLI: Questions Natural Language Inference \cite{QNLI-SQuAD2} dataset is a modified version of the SQuAD dataset \cite{SQuAD}. It contains over $100k$ sentence/context pairs where the aim is to determine if the context contains an answer to the question.
    \item RTE: Recognizing  Textual  Entailment \cite{RTE} is similar to MNLI, where the aim is to predict the type of entailment between a paragraph and a sentence, entailment, contradiction, and unknown being the choices. 
    \item WNLI: Winograd Natural Language Inference \cite{WNLI} dataset also concerns natural language inference similar to the MNLI and the RTE datasets.
\end{itemize}

Stanford Question Answering Dataset (SQuAD 1.1 \cite{SQuAD} and SQuAD 2.0 \cite{squad2}) is a reading comprehension dataset that is composed of Wikipedia articles and questions related to them. The aim is to find the text segment that answers the related question. There are $150k$ questions $50k$ of which is unanswerable from the given context article. Any model built for this task should also determine whether the question is answerable or not in addition to answering the questions.

RACE dataset \cite{race} is also a dataset for reading comprehension taken from the English exams for middle and high school Chinese students. The aim is to find the correct answer to the question about a specific text passage among the choices. There are approximately $28k$ passages and $100k$ questions.

One can find the links of the datasets in Appendix \ref{app-details}. The leaderboards of current state-ot-the-art can be tracked either from the respective websites or from the ACL Wiki website (\url{https://aclweb.org/aclwiki/State_of_the_art}). The reader can refer to \citet{surveyevaluation} for comparisons, advantages, and disadvantages of the evaluation methods of word embedding models.

\subsection{Results}

\begin{table}[t]
	\centering
	\caption{Word embedding models' performances in similarity tasks (in chronological order).}
	\label{tab:performance_similarity}
	\adjustbox{width=\columnwidth}{
	\begin{tabular}{c|c|c|ccccc|c|c|c|c|c|c}
		\hline
		 Model & Dim. & WS-353  & \multicolumn{5}{c|}{SCWS ($\rho \times 100$)} & RG-65 & MEN & YP-130 & RW & MC-30 & Simlex-999\\
		  & & ($\rho \times 100$) & avgSim & avgSimC & globalSim & localSim & MaxSimC & ($\rho \times 100$) & ($\rho \times 100$) & ($\rho \times 100$) & ($\rho \times 100$) & ($\rho \times 100$) & ($\rho \times 100$)\\
		\hline \hline
		
		HLBL \cite{mnih2008} & 100 & 33.2\footnotemark[3] & - & - & - & - & - & - & - & - & - & - & -\\
		
		C\&W \cite{Collobert2008} & 50 & 29.5\footnotemark[3] & - & - & 57.0\footnotemark[8] & - & - & 48.0\footnotemark[7]  & 57.0\footnotemark[7] & - & - & - & -\\
		C\&W \cite{Collobert2008} & 50 & 49.8\footnotemark[3] & - & - & - & - & - & - & - & - & - & - & -\\
	
		R\&M \cite{reisinger2010} & - & 73.4\footnotemark[3] & 60.4\footnotemark[3] & 60.5\footnotemark[3] & 62.5\footnotemark[8] & - & 60.4\footnotemark[8] & - & - & - & - & - & -\\
		
		RNNLM \cite{mikolov2010} & 640 & - & - & - & - & - & - & - & - & - & - & - & -\\
		
		Huang et al. \cite{huang2012}  & 50 & 71.3 & 62.8 & 65.7 & 58.6\footnotemark[2] & 26.1\footnotemark[2] & - & - & - & - & - & - & -\\

		CBOW \cite{mikolov2013a} & 400 & 69.4\footnotemark[7] & 64.2\footnotemark[7] & - & - & - & - & 73.2\footnotemark[7] & 66.5\footnotemark[7] & 34.3\footnotemark[7] & - & - & -\\ 
		
		Skip-Gram \cite{mikolov2013a} & 100 & 58.9\footnotemark[5] & - & - & - & - & - & -  & - & - & - & - & -\\
		
		Skip-Gram \cite{mikolov2013b} & 300 & 70.4\footnotemark[9] & 66.6\footnotemark[6] & 66.6\footnotemark[6] & 65.2\footnotemark[9] & - & - & - & - & - & - & - & -\\ 
		
		Skip-Gram \cite{mikolov2013b} & 256 & 66.7\footnotemark[1] & - & - & - & - & - & - & 55.7\footnotemark[1] & - & 38.8\footnotemark[1] & - & -\\
		
		Luong et al.\cite{luong2013} & 50 & 64.6 & - & - & 48.5 & - & - & 65.4 & - & - & 34.4 & 71.7 & -\\
		
		CLBL \cite{botha2014} & - & 39.0 & - & - & - & - & - & 41.0 & - & - & 30.0 & - & -\\

        Tian et al. \cite{tian2014} & 50 & - & - & 65.4 & - & - & 63.6 & - & - & - & - & - & -\\
        
        Qiu et al. \cite{qiu2014} & 200 & 65.2 & - & - & 53.4 & - & - & 67.4 & - & - & 32.9 & 81.6 & -\\
        
        MSSG \cite{neelakantan2014} & 300 & 70.9 & 67.3 & 69.1 & 65.5 & 59.8 & - & - & - & - & - & - & -\\
        
        Chen et al. \cite{chen2014} & 200 & - & 66.2 & 68.9 & 64.2 & - & - & - & - & - & - & - & -\\
        
        GloVe \cite{pennington2014} & 300 & 75.9 & - & - & 59.6 & - & - & 82.9 & - & - & 47.8 & 83.6 & 41.0\footnotemark[15]\\
        
        Guo et al. \cite{guo2014} & 50 & - & 49.3 & - & - & - & 55.4 & - & - & - & - & - & -\\ 
        
        KNET \cite{cui2015} & 100 & 66.1 & - & - & - & - & - & - & - & - & 39.3 & - & -\\

		CNN-VMSSG \cite{Chen2015} & 300 & - & 65.7 & 66.4 & 66.3 & \textbf{61.1} & - & -  & - & - & - & - & -\\
		
		AutoExtend \cite{rothe2015} & 300 & - & \textbf{68.9} & \textbf{69.8} & - & - & - & -  & - & - & - & - & -\\

        SenseEmbed \cite{iacobacci2015} & 400 & \textbf{77.9} & 62.4 & - & - & - & - & \textbf{89.4} & 80.5 & \textbf{73.4} & - & - & -\\
        
        TWE-1 \cite{liu2015a} & 400 & - & - & 68.1 & - & - & \textbf{67.3} & - & - & - & - & - & -\\
        
        Jauhar et al. \cite{jauhar2015} & 80 & 63.9 & - & - & 65.7 & - & - & 73.4 & 64.6 & - & - & 75.8 & -\\
        
        SAMS \cite{cheng2015} & 300 & - & 62.5 & - & 59.9 & 58.5 & - & - & - & - & - & - & -\\
        
        SWE \cite{liu2015b} & 300 & 72.8 & - & - & - & - & - & - & - & - & - & - & -\\
        
        Soricut and Och \cite{soricut2015} & 500 & 71.2 & - & - & - & - & - & 75.1 & - & - & 41.8 & - & -\\
		
		Cotterell et al. \cite{cotterell2016} & 100 & 58.9 & - & - & - & - & - & - & - & - & - & - & -\\
		
		char2vec \cite{cao2016} & 256 & 34.5 & - & - & - & - & - & - & 32.2 & - & 28.2 & - & -\\
		
		Bojanowski et al. \cite{bojanowski2016} & 300 & 71.0 & - & - & - & - & - & - & - & - & 47.0 & - & -\\
		
		Yin and Schütze \cite{yin2016} & 200 & 76.0 & - & - & - & - & - & - & \textbf{82.5} & - & \textbf{61.6} & 85.7 & 48.5\\
		
		dLCE \cite{nguyen2016} & 500 & - & - & - & - & - & - & - & - & - & - & - & \textbf{59.0}\\
		
		Ngram2vec \cite{zhao2017} & 300 & - & - & - & - & - & - & - & 76.0 & - & 44.6 & - & 42.1\\
		
		MSWE \cite{nguyenDQ2017} & 300 & 72.4 & 66.7 & 66.7 & \textbf{66.8} & - & - & - & 76.4 & - & 35.6 & - & 39.2\\
		
		Dict2vec \cite{tissier2017} & 300 & 75.6 & - & - & - & - & - & 87.5 & 75.6 & 64.6 & 48.2 & \textbf{86.0}  & -\\
		
		LMM \cite{xu2018} & 200 & 61.5 & - & - & 63.0 & - & - & 63.1 & - & - & 43.1 & -  & -\\
		
		LSTMEmbed \cite{iacobacci2019} & 400 & 61.2 & - & - & - & - & - & - & - & - & - & -  & -\\

    
		\hline
	\end{tabular}
	}
\end{table}

In this section, we report the results obtained by the models examined in this survey on aforementioned datasets. In Tables \ref{tab:performance_similarity},\ref{tab:performance_analogy},\ref{tab:performance_synonym}, and \ref{tab:performance_downstream}, the results in similarity, analogy, synonym selection, and downstream tasks are given respectively. 

While reporting the results, we follow a few criteria to make it as fair and simple as possible:
\begin{itemize}
    \item Unless noted otherwise, all of the results are taken from the original papers. (The results taken from other sources are marked with numbered superscripts. See Appendix \ref{app-details} for details.)
    \item If more than one paper report results on the same model, we take the one in the original paper.
    \item If the author(s) provide several variations of a model, we report only the one with the best score.
\end{itemize}

Although some of the differences in performances of word representations are due to the models themselves, it should be noted that the size of the datasets that the models are trained on can be different, therefore, can affect the fairness of comparison.

Table \ref{tab:performance_similarity} shows word embedding models' performances in similarity tasks. SenseEmbed \cite{iacobacci2015} is the best performing model in WS-353, RG-65, and YP-130 datasets according to the reported results. \citet{yin2016} has superior performance in the datasets of MEN and RW, while Dict2vec \cite{tissier2017} outperforms others on MC-30. In SCWS, AutoExtend \cite{rothe2015} gives the highest correlation coefficient scores. In general, GloVe \cite{pennington2014}, SenseEmbed \cite{iacobacci2015}, \citet{yin2016}, and Dict2vec \cite{tissier2017} perform well on similarity datasets.

SenseEmbed's \cite{iacobacci2015} success can be attributed to its capability to disambiguate senses by being trained on sense-tagged corpora. Glove \cite{pennington2014} is generally robust as it's a mixture of global co-occurrence and local context-based methods. When it comes to \citet{yin2016}, it is an ensemble of existing embeddings, including Glove, which produces better representations for OOV words due to its ensemble nature. Thus, it has good coverage of words in similarity datasets. Dict2vec's \cite{tissier2017} performance proves the effectiveness of positive sampling over word2vec \cite{mikolov2013b}.

Word embedding models' performances are tested on Google Analogy Task that includes both syntactic and semantic analogies (Table \ref{tab:performance_analogy}). The best accuracy scores are obtained by \citet{yin2016} in this category. Glove\cite{pennington2014} follows it as the second-best performing model.
Results in the Google Analogy task can be interpreted much as those in similarity tasks.

\begin{table}[t]
	\centering
	\caption{Word embedding models' performances in analogy task (in chronological order).}
	\label{tab:performance_analogy}
	\adjustbox{width=0.5\columnwidth}{
	\begin{tabular}{c|c|ccc}
		\hline
		 Model & Dimension & \multicolumn{3}{|c}{Google Analogy Task (acc. \%)}\\
		  & & Syntactic & Semantic & Total \\
		\hline \hline
		
		C\&W \cite{Collobert2008} & 50 & 9.3\footnotemark[4] & 12.3\footnotemark[4] & 11.0\footnotemark[4] \\
		
		RNNLM \cite{mikolov2010} & 640 & 8.6\footnotemark[4] & 36.5\footnotemark[4] & 24.6\footnotemark[4] \\
		
		CBOW \cite{mikolov2013a} & 1000 & 57.3 & 68.9 & 63.7 \\ 
				
		Skip-Gram \cite{mikolov2013a} & 1000 & 66.1 & 65.1 & 65.6 \\ 
		
		Skip-Gram \cite{mikolov2013b} & 100 & 36.4\footnotemark[13] & 28.0\footnotemark[13] & 32.6\footnotemark[13] \\
		
		Skip-Gram \cite{mikolov2013b} & 300 & 61.0 & 61.0 & 61.0 \\ 
		
		Skip-Gram \cite{mikolov2013b} & 256 & 51.3\footnotemark[1] & 33.9\footnotemark[1] & 43.6\footnotemark[1] \\
	
		ivLBL \cite{mnih2013} & 100 & 46.1 & 40.0 & 43.3\\
		
		ivLBL \cite{mnih2013} & 300 & 63.0 & 65.2 & 64.0\\
		
		vLBL \cite{mnih2013} & 300 & 64.8 & 54.0 & 60.0\\
		
		vLBL \cite{mnih2013} & 600 & 67.1 & 60.5 & 64.1\\
		
		Qiu et al. \cite{qiu2014} & 200 & 58.4 & 25.0 & 43.3\\
		
		MSSG \cite{neelakantan2014} & 300 & - & - & 64.0\footnotemark[10] \\
		
		GloVe \cite{pennington2014} & 300 & 69.3 & 81.9 & 75.0 \\
		
        KNET \cite{cui2015} & 100 & 46.9 & 24.9 & 36.3 \\
		
		char2vec \cite{cao2016} & 256 & 52.5 & 2.5 & 35.5 \\
		
		Bojanowski et al. \cite{bojanowski2016} & 300 & 74.9 & 77.8 & - \\
		
		Yin and Schütze \cite{yin2016} & 200 & \textbf{76.3} & \textbf{92.5} & \textbf{77.0} \\
		
		Ngram2vec \cite{zhao2017} & 300 & 71.0 & 74.2 & 72.5 \\
		
		MSWE \cite{nguyenDQ2017} & 50 & - & - & 69.9 \\
		
		LMM \cite{xu2018} & 200 & 20.4 & - & - \\
		
		\hline
	\end{tabular}
	}
\end{table}

In synonym-selection tasks, three models' (Skip-Gram \cite{mikolov2013b}, Jauhar et al. \cite{jauhar2015}, SWE \cite{liu2015b})  results are reported (Table \ref{tab:performance_synonym}). In ESL-50 and RD-300 datasets, the only model with the reported performance is Jauhar et al. \cite{jauhar2015}. In TOEFL-80, SWE\cite{liu2015b} outperforms the others. Here, SWE's success can be explained by its synonym-antonym rule in learning word embeddings.

In Table \ref{tab:performance_downstream}, word embedding models' performances on downstream tasks are provided. In GLUE benchmark, CBOW \cite{mikolov2013a}, BiLSTM+Cove+Attn \cite{McCann2017}, and BiLSTM+Elmo+Attn \cite{peters2018} are behind human baselines except for the task of QQP. In QQP, CBOW is still underperforming but BiLSTM+Cove+Attn \cite{McCann2017} and BiLSTM+Elmo+Attn\cite{peters2018} are superior to human performance. 

As for the original BERT\cite{devlin2019} and its variants, in the tasks of MRPC, QQP, QNLI, they consistently outperform human baselines. In SST-2, MNLI, RTE, and WNLI, human performance is better. In STS-B, the only model with superior performance to humans is ALBERT\cite{albert}, In CoLA, and the tasks of question answering (SQuAD 2.0), and reading comprehension (RACE), starting from XLNET\cite{xlnet} better performances over human are observed. GPT-3 \cite{gpt-3} is promising with its language model meta-learner idea and gives its best performance in the Few-Shot setting. Although it is behind the state-of-the-art by a large margin in GLUE benchmark, in RTE its score is beyond CBOW \cite{mikolov2013a}, BiLSTM+Cove+Attn \cite{McCann2017}, and BiLSTM+Elmo+Attn \cite{peters2018}.

Table \ref{tab:performance_downstream} proves the success of contextual representations, especially the transformer-based models (BERT \cite{devlin2019} and its successors), by going beyond human performance in most of the downstream tasks. However, it can be said that in natural language inference tasks such as MNLI, WNLI, and RTE, these probabilistic language representations still have some limitations in meeting causal inference requirements.  

\begin{table}[t]
	\centering
	\caption{Word embedding models' performances in synonym selection tasks (in chronological order).}
	\label{tab:performance_synonym}
	\adjustbox{width=0.5\columnwidth}{
	\begin{tabular}{c|c|c|c|c}
		\hline
		 Model & Dimension & ESL-50 (\%) & TOEFL-80 (\%) & RD-300 (\%) \\
		\hline \hline
		
		Skip-Gram \cite{mikolov2013b} & 300 & - & 83.7\footnotemark[11] & - \\
		
		Skip-Gram \cite{mikolov2013b} & 400 & 62.0\footnotemark[14] & 87.0\footnotemark[14] & - \\
		
		GloVe \cite{pennington2014} & 300 & 60.0\footnotemark[14] & 88.7\footnotemark[14] & - \\
		
		MSSG \cite{neelakantan2014} & 300 & 57.1\footnotemark[14] & 78.3\footnotemark[14] & - \\
        
        Jauhar et al. \cite{jauhar2015} & 80 & 63.6 & 73.3 & 66.7 \\
		
        Jauhar et al. \cite{jauhar2015} & 80 & \textbf{73.3}\footnotemark[14] & 80.0\footnotemark[14] & - \\
        
        Li and Jurafsky \cite{li2015} & 300 & 50.0\footnotemark[14] & 82.6\footnotemark[14] & - \\
        
        SWE \cite{liu2015b} & 300 & - & 88.7 & - \\
        
        LSTMEmbed \cite{iacobacci2019} & 400 & 72.0 & \textbf{92.5} & - \\

		\hline
	\end{tabular}
	}
\end{table}

\begin{table}[t]
	\centering
	\caption{Word embedding models' performances in downstream tasks.}
	\label{tab:performance_downstream}
	\adjustbox{width=\columnwidth}{
	\begin{tabular}{c|c|c|c|c|c|c|c|c|c||c|c}
		\hline
		 Model & CoLA & SST-2 & MRPC & STS-B & QQP & MNLI & QNLI & RTE & WNLI & SQuAD 2.0 & RACE\\
		                & (mcc) & (\%) & (F1) & ($\rho \times 100$) & (F1) & m/mm (\%/\%) & (\%) & (\%) & (\%) & (F1) & (\%) \\
		\hline \hline
		  CBOW \cite{mikolov2013a} & 0.0 & 80.0 & 81.5 & 58.7 & 51.4 & 56.0/56.4 & 72.1 & 54.1 & 62.3 &  &  \\ 
		  BiLSTM+Cove+Attn \cite{McCann2017} & 8.3 & 80.7 & 80.0 & 68.4 & 60.5 & 68.1/68.6 & 72.9 & 56.0 & 18.3 & - & - \\
		  BiLSTM+Elmo+Attn \cite{peters2018} & 33.6 & 90.4 & 84.4 & 72.3 & 63.1 & 74.1/74.5 & 79.8 & 58.9 & 65.1 & - & - \\
		  GLUE Human Baselines & 66.4 & 97.8 & 86.3 & 92.6 & 59.5 & 92.0/92.8 & 91.2 & 93.6 & 95.9 & - & - \\
		  SQuAD Human Baselines \cite{squad2} & - & - & - & - & - & - & - & - & - & 89.4 & - \\
		  Turkers \cite{race} & - & - & - & - & - & - & - & - & - & - & 73.3 \\
		  \hline  
		    
		  BERT \cite{devlin2019} & 60.5 & 94.9 & 89.3 & 86.5 & 72.1 & 86.7/85.9 & 91.1 & 70.1 & 65.1 & 89.1\footnotemark[12] & 72.0\footnotemark[12] \\ 
		  
		  ERNIE 2.0 \cite{ernie2} & 63.5 & 95.6 & 90.2 & 90.6 & 73.8 & 88.7/88.8 & 94.6 & 80.2 & 67.8 & - & - \\
		  
		  XLNet \cite{xlnet} (ensemble) & 67.8 & 96.8 & 92.9 & 91.6 & \textbf{74.7} & 90.2/89.7 & 98.6 & 86.3 & 90.4 & 89.1\footnotemark[12] & 81.8\footnotemark[12] \\
		  
		  RoBERTa \cite{roberta} (ensemble) & 67.8 & 96.7 & 92.3 & 91.9 & 74.3 & 90.8/90.2 & 98.9 & 88.2 & 89.0 & 89.8\footnotemark[12] & 83.2\footnotemark[12] \\
		  
		  ALBERT \cite{albert} & \textbf{71.4} & 96.9 & 90.9 & \textbf{93.0} & - & 90.8 & 95.3 & 89.2 & - & 90.9 & 86.5 \\ 

		  ALBERT \cite{albert} (ensemble) & 69.1 & \textbf{97.1} & \textbf{93.4} & 92.5 & 74.2 & \textbf{91.3}/\textbf{91.0} & \textbf{99.2} & \textbf{89.2} & \textbf{91.8} & \textbf{92.2} & \textbf{89.4} \\ 

		  GPT-3 Few-Shot \cite{gpt-3} - & - & - & - & - & - & - & - & 69.0 & - & 69.8 & 45 \\ 

		\hline
	\end{tabular}
	}
\end{table}

\section{Conclusion}

Human-level language understanding is one of the oldest challenges in computer science. Many scientific works have been dedicated to finding good representations for semantic units (words, morphemes, characters) in languages since it is preliminary for all downstream tasks in NLP. Most of these studies use the distributional hypothesis, where the meaning of a word is measured from its neighboring words.

Distributed representation through a neural network is intuitive in that it resembles human mind's representation of concepts. Beyond that, pre-trained language models' knowledge has been transferred to fine-tuned task-specific models, which introduced a boost in performance. To summarize, neural language models with their updated weights as well as learned representations in their layers have become a source of knowledge.

From the release of early word embeddings to current contextual representations, the area of semantics has experienced a transformation, which becomes evident by substantial performance improvements in all NLP tasks. The idea of pre-training a language model then fine-tuning it on a downstream task has become a \textit{de facto} standard in almost all subfields of NLP.

Recently, contextual models, such as BERT and its variants, showed great success in downstream NLP tasks using masked language modeling and transformer structures. They have become state-of-the-art word embeddings and obtained human-level results on some of the downstream tasks.

Over the last few years, there has been an increase in the studies that consider experiential (visual) information by building multi-modal language models and representations \citep{sezerer2021, vilbert, SAN}. The idea of multi-modal language modeling is based on human language acquisition, where learning starts with concrete concepts through images early on (As pointed out by the "pointing phase" in children \citep{pointing1, pointing2}) and then continues with learning abstract ideas through the text \citep{vigliocco2009, andrews2009, Griffiths2007}. Fueled by the success of text-based language models and advancements in cognitive psychology, perhaps this type of multi-modal language modeling can be the next goal to tackle in the future.

\begin{acks}
We want to thank Tuğkan Tuğlular for his valuable comments and feedback in the development of this survey.
\end{acks}

\nocite{Griffiths2007}
\nocite{vigliocco2009}
\nocite{andrews2009}
\nocite{schutze1998}

\bibliographystyle{ACM-Reference-Format}
\bibliography{bibliography}

\appendix

\section{Reading Guide for Beginners}\label{app-guide}
\begin{figure*}[h]
    \centering
	\includegraphics[width=0.85\textwidth]{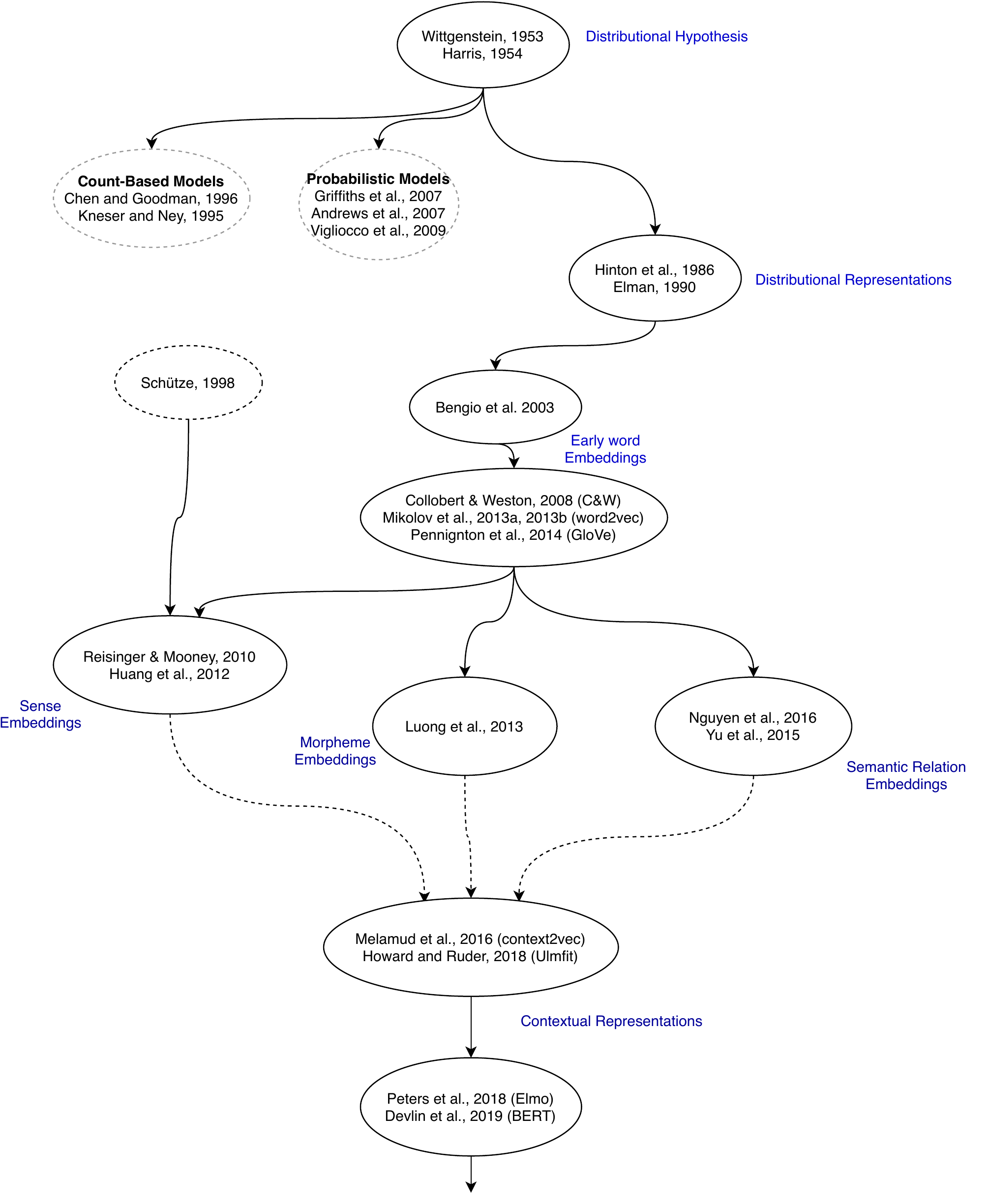}
	\caption{Evolution of neural word embeddings.}
	\label{reading_guide}
\end{figure*}

In Figure \ref{reading_guide}, the milestone papers of each subtopic are listed. It is not an exhaustive list created to serve as a starting point for researchers who are not familiar with the subject. For the details of each subfield, readers should refer to the corresponding chapters in the survey. Dashed sets represent non-neural models, which are not the subject of this survey.

\section{Details on Datasets and Results}\label{app-details}
\subsection{Datasets}
 
\begin{itemize}
    \item WS-353: \url{http://gabrilovich.com/resources/data/wordsim353/wordsim353.zip}
    
    \item SCWS: \url{http://www-nlp.stanford.edu/~ehhuang/SCWS.zip}
    
    \item RG-65: \textit{There are no formal links to this dataset}
    
    \item MC-30: \textit{There are no formal links to this dataset}
    
    \item MEN: \url{https://staff.fnwi.uva.nl/e.bruni/MEN}
    
    \item YP-130: \url{https://www.researchgate.net/publication/257946337_Verb_similarity_on_the_taxonomy_of_WordNet_-_dataset/link/02e7e5266fe99269cc000000/download}
    
    \item RW: \url{http://www-nlp.stanford.edu/~lmthang/morphoNLM/rw.zip}
    
    \item Simlex-999: \url{https://fh295.github.io/simlex.html}
    
    \item Google Analogy Task: \url{http://download.tensorflow.org/data/questions-words.txt}
    
    \item ESL-50: \url{https://www.apperceptual.com/home} (personal communication)
    
    \item TOEFL-80: \url{http://lsa.colorado.edu/mail_sub.html} (personal communication)
    
    \item RD-300: \url{https://arxiv.org/ftp/arxiv/papers/1204/1204.0140.pdf} (Appendix K; also contains TOEFL-80 and ESL-50)    
    
    \item Glue Benchmark (CoLA, SST-2, MRPC, STS-B, QQP, MNLI, QNLI, RTE, WNLI):\\ \url{https://gluebenchmark.com/tasks}
     
    \item Stanford Question Answering Dataset (SQuAD 1.1 \cite{SQuAD} and SQuAD 2.0 \cite{squad2}):\\ \url{https://rajpurkar.github.io/SQuAD-explorer/}
    
    \item RACE dataset: \url{http://www.cs.cmu.edu/~glai1/data/race/}  
    
\end{itemize}

\subsection{Results}
You can find the sources of experimental results here. Each number corresponds to the numbered superscripts used in the tables:\\
1: reported in \cite{cao2016}\\
2: reported in \cite{Chen2015}\\
3: reported in \cite{huang2012}\\
4: reported in \cite{mikolov2013a}\\
5: reported in \cite{cotterell2016}\\
6: reported in \cite{rothe2015}\\
7: reported in \cite{iacobacci2015}\\
8: reported in \cite{liu2015a}\\
9: reported in \cite{neelakantan2014}\\
10: reported in \cite{nguyenDQ2017}\\
11: reported in \cite{liu2015b}\\
12: reported in \cite{albert}\\
13: reported in \cite{mnih2013}\\
14: reported in \cite{iacobacci2019}\\
15: reported in \cite{mrksic2016}\\

\end{document}